\documentclass{article}
\usepackage{PRIMEarxiv}

\usepackage{graphicx}
\usepackage[english]{babel}
\usepackage{times} 
\usepackage{url}
\usepackage{float}
\usepackage{verbatim}
\usepackage{mwe}
\usepackage{color}
\usepackage{hyperref}
\usepackage{amsmath}
\usepackage{bm}
\usepackage{multirow}
\usepackage[numbers]{natbib}

\usepackage{times}
\usepackage[utf8]{inputenc}

\usepackage{tabularx} 
\usepackage{ulem}
\usepackage[ruled,vlined,linesnumbered,noend]{algorithm2e}

\usepackage{caption}
\usepackage{subcaption}

\usepackage[misc]{ifsym}

\usepackage{xcolor}
\usepackage[colorinlistoftodos,prependcaption,textsize=tiny]{todonotes}

\emergencystretch=\maxdimen
\title{MisRoBÆRTa: Transformers versus Misinformation}

\author{
    Ciprian-Octavian~Truică$^{1,2}$,
    Elena-Simona~Apostol$^{1,2}$ \\
    $^1$ Department of Information Technology, Uppsala University, Uppsala, Sweden\\
    $^2$ Computer Science and Engineering Department, Faculty of Automatic Control and Computers, \\ University Politehnica of Bucharest, Bucharest, Romania \\
  \texttt{ciprian.truica@upb.ro, elena.apostol@upb.ro}
}

\begin{document}

\maketitle              

\begin{abstract}
Misinformation is considered a threat to our democratic values and principles. 
The spread of such content on social media polarizes society and undermines public discourse by distorting public perceptions and generating social unrest while lacking the rigor of traditional journalism.
Transformers and transfer learning proved to be state-of-the-art methods for multiple well-known natural language processing tasks.
In this paper, we propose MisRoBÆRTa, a novel transformer-based deep neural ensemble architecture for misinformation detection.
MisRoBÆRTa takes advantage of two state-of-the art transformers, i.e., BART and RoBERTa, to improve the performance of discriminating between real news and different types of fake news.
We also benchmarked and evaluated the performances of multiple transformers on the task of misinformation detection.
For training and testing, we used a large real-world news articles dataset (i.e., 100,000 records) labeled with 10 classes, thus addressing two shortcomings in the current research:
(\textit{1}) increasing the size of the dataset from small to large, and
(\textit{2}) moving the focus of fake news detection from binary classification to multi-class classification.
For this dataset, we manually verified the content of the news articles to ensure that they were correctly labeled.
The experimental results show that the accuracy of transformers on the misinformation detection problem was significantly influenced by the method employed to learn the context, dataset size, and vocabulary dimension.
We observe empirically that the best accuracy performance among the
classification models that use only one transformer is obtained by BART, while DistilRoBERTa obtains the best accuracy in the least amount of time required for fine-tuning and training.
However, the proposed MisRoBÆRTa outperforms the other transformer models in the task of misinformation detection.
To arrive at this conclusion, we performed ample ablation and sensitivity testing with MisRoBÆRTa on two datasets. 
\end{abstract}

\section{Introduction}
The digital age has seen the emergence of new mass media paradigms for information distribution, substantially different from classical mass media.
With new digital-enabled mass media, the communication process is centered around the user, while multimedia content is the new identity of news.
Thus, the media landscape has shifted from mass media to personalized social media with these new paradigms.
Along with the advantages this progress brings, it aggravates the risk of fake news~\cite{Ruths2019}, potentially with  detrimental consequences to society, by facilitating the spread of misinformation in the form of propaganda, conspiracy theories, political bias, etc.

Misinformation consists of news articles that are intentionally and verifiably false or inaccurate and which could mislead readers by presenting alleged, imaginary, or~false facts about social, economic, and political subjects of interest~\cite{Shu2018,Ilie2021}.
However, current trends and the accessibility of technology render this proneness even more potentially damaging than it has been historically, having dire consequences to the community and society at large~\cite{Zannettou2019}.
Its spread influences communities and creates public polarization regarding elections~\cite{Bovet2019}, medical practices~\cite{MarcoFranco2021}, etc.
Furthermore, the practices of misinforming society are been condemned by politicians and businessmen alike, being considered real threats that undermine democracy and public health~\cite{ECDisinformation2021}; we believe that tools and methods for accurately detecting misinformation are of high relevance to the research community and society at~large.

The main motivation for this paper is to evaluate the performance of different transformers for the task of misinformation detection in order to propose a novel deep learning transformer-based model, i.e., MisRoBÆRTa.
We used a large real-world news dataset consisting of 100,000 news articles, labeled using 10 different classes, i.e.,~fake news, satire, extreme bias, hate news, etc.
By employing such a dataset, we also address two shortcomings in the current~research:
\begin{itemize}
    \item[(\textit{1})] Increasing the size of the dataset from small to large; and~
    \item[(\textit{2})] Moving the focus of fake news detection from binary to multi-class classification.
\end{itemize}
The experimental results show that the accuracy of transformers on the multi-class classification task is not influenced by the size of the network, but rather by the method employed to learn the context, and~the size of both the dataset and vocabulary.
We also include in our benchmark the state-of-the-art FakeBERT model~\cite{Kaliyar2021}.

The research questions for our benchmark are as follows:
\begin{itemize}
    \item[(\textit{Q1})] Which transformer obtains the overall best accuracy for the task of multi-class classification of misinformation?
    \item[(\textit{Q2})] Which transformer obtains the overall best runtime performance without the model having a significant decrease in accuracy?
    \item[(\textit{Q3})] Can there be a balance between accuracy and runtime required for fine-tuning and training?
\end{itemize}

Our contributions are as follows:
\begin{enumerate}
    \item[(\textit{1})] We propose MisRoBÆRTa, a~new transformer-based deep learning architecture for misinformation detection;
    
    \item[(\textit{2})] We conducted an in-depth analysis of the current state-of-the-art deep learning, word embeddings, and~transformer-based models for the task of fake news and misinformation detection;

    \item[(\textit{3})] We propose a new balanced dataset for misinformation containing 100,000 articles extracted from the FakeNewsCorpus, where, for each news article, we (1) manually verified its content to make sure that it was correctly labeled; (2)verified the URLs to point to the correct article by matching the titles and authors; 

    \item[(\textit{4})] We performed an in-depth analysis of the proposed dataset;

    \item[(\textit{5})] We extended the SimpleTransformers package with a multi-class implementation for~BART;

    \item[(\textit{6})] We performed a detailed benchmark analysis using multiple transformers and transfer learning on the task of misinformation and compared the results with the state-of-the-art model FakeBERT~\cite{Kaliyar2021}.
\end{enumerate}

This paper is structured as follows.
Section~\ref{sec:related_work} discusses the current research and methods related to misinformation and fake news detection.
Section~\ref{sec:methodology} introduces the transformers employed in this study as well as the neural network used for classification.
Section~\ref{sec:results} presents the datasets and analyzes the experimental results.
Section~\ref{sec:discussion} discusses our findings and hints at current challenges that we identify for the task of misinformation detection.
Section~\ref{sec:conclusions} presents the conclusions and outlines future directions.

\section{Related~Work}\label{sec:related_work}
The misinformation and fake news detection problem has been tackled in the literature using different perspectives, from the linguistic approach to more sophisticated natural language processing (NLP) models.
Whereas, the linguistic approach attempts to learn language cues that are hard to monitor (e.g., frequencies and patterns of pronouns, conjunctions, and negative emotion word usage~\cite{Conroy2015}), the NLP approaches build models using content-based features for fake news detection~\cite{Gravanis2019}.

In article~\cite{Kaliyar2020}, a~deep convolutional-based GloVe-enabled model (FNDNet) is used for fake news detection.
Although the experimental setup for the tested networks uses small epoch numbers, FNDNet shows  promising results compared to other deep learning architectures, i.e., classical  convolutional neural network (CNN) and long short-term memory (LSTM), both trained using GloVe~\cite{Pennington2014}.
Another CNN-based architecture that shows promising results is MCNN-TFW (multiple-level convolutional neural network-based)~\cite{Li2019}. 
MCNN-TFW uses semantics features that are determined using a pre-trained Word2Vec~\cite{Mikolov2013}.

Attention-based models are also good candidates for fake news detection.
In~\cite{Mishra2019}, a new hierarchical attention network is used for learning GloVe embeddings for different latent aspects of news articles. 
The solution uses attention weights to support or refute the analyzed claims.
The authors emphasize the importance of aspect embeddings for false news prediction.
In \citet{Trueman2021}, the authors present an attention-based convolutional bidirectional LSTM approach to detect fake news.
The news articles are classified using six multilevel categories ranging from ``true" to ``pants on fire". 
The authors use the LIAR dataset consisting of 12,836 statements for training, validation, and testing.

Other current approaches use advanced pre-trained model-based detection, i.e., transformers~\cite{Liu2019,Liu2019BertFN,Kurasinski2020,Jwa2019}, or~new vectorization techniques, i.e., CLDF~\cite{Mersinias2020}.
Among the transformers, bidirectional encoder representations from transformers (BERT) (Section~\ref{sec:methodology}) has been used the most for fake news detection.

\citet{Liu2019BertFN} proposed a two-stage BERT-based model for fake news detection.
Unlike simple BERT, the authors utilized all hidden states and applied attention mechanisms to calculate weights.
The experiments were done using the LIAR dataset containing 12,836 short news statements.
Being a relatively small dataset with little information, the proposed model and the compared ones have trouble learning correctly.
Furthermore, current research proves that BERT does not provide accurate models if it is fine-tuned with a small dataset or short statements~\cite{Shaar2020}. 

A comparison in performance between BERT and a deep-learning architecture using CNN and BiLSTM layers is presented in~\cite{Kurasinski2020}. 
This article also offers an analysis of the attention coverage of texts for the two models.
Although two small ($\sim$1,000 articles) subsets from the FakeNewsCorpus dataset are used, the~classification is not multi-class, as~the authors transformed the 11 categories into 2 classes, i.e.,~real and fake.
The results show that, on the two small datasets, BiLSTM-CNN architecture outperforms or has the same accuracy as BERT; thus, strengthening the statement from~\cite{Shaar2020}.

In~\cite{Kula2020}, a BERT-RNN architecture uses a relatively big balanced dataset and obtains very good results on a dataset labeled with different levels of veracity i.e., different degrees of truth and falsity.
This solution has a hybrid architecture connecting BERT word embeddings with a recurrent neural network (RNN). 
The RNN network was used to obtain document embeddings. 
The major issue with this solution is that the results are not explained or discussed, and there is no comparison with other embedding models or deep neural networks.

Another hybrid BERT-based solution is FakeBERT~\cite{Kaliyar2021}.
FakeBERT uses the vectors generated after word embedding from BERT as input to several 1D-CNN layers.
The authors compared their solution with other deep learning solutions, considering a dataset from Kaggle with binary labels.
The same dataset is used in FNDNet~\cite{Kaliyar2020}.
By analyzing these two articles written by the same main authors, we can conclude that FakeBERT has slightly better accuracy than FNDNet.
Thus, we decided to add FakeBERT to our benchmark.

Other transformers did not get as much attention as BERT, but~still, there are some articles that tackle this subject.
A weak social supervision RoBERTa-based solution to detect fake news from limited annotated data is proposed in~\cite{Shu2020RoBERTa}. 
RoBERTa is an optimized method that improves BERT (Section~\ref{sec:methodology}).
Their model outperforms other CNN or adversarial neural network-based solutions.
In other papers, we found fake news detection solutions based on XLNet, an extension of BERT.
In \citet{Gautam2021}, the authors combined topic information with an XLNet model.
The authors compared their solution (XLNet + LDA) with basic XLNet, and also with BERT on a relatively small annotated dataset ($\sim$10~K). 
To achieve an increase in time performance, some solutions use compact language models, such as  a lite BERT (ALBERT)~\cite{Lan2020}.
FakeFinder~\cite{Tian2020} is an ALBERT-based solution for mobile devices that offers real-time fake news detection from Twitter. 
This small ALBERT-based model achieves comparable F1 scores with a BERT model, although slightly smaller when tested on a dataset with $\sim$2,500 tweets and $\sim$117,000 comments.

To the best of our knowledge, there are almost no benchmarks on misinformation detection using transformers, a multi-class task.
The only (found) article that presents a more in-depth benchmark on this subject is~\cite{Khan2021benchmark}.
The authors only considered binary classification for five pre-trained models (i.e., BERT, RoBERTa, DistilBERT, ELECTRA, ELMo~\cite{Peters2018deep}). 
The best performance was obtained by RoBERTa, but it is not clear if they used fine-tuning.

After the state-of-the-art analysis, we noticed the following: none of these methods attempted to detect different types of misinformation, e.g., clickbait, extreme bias, but rather focused on the veracity of the articles, binary or multilevel~\cite{Kula2020}, i.e.,~different degrees of truth and falsity.
Moreover, the majority of the solutions used relatively small datasets.

\section{Methodology}\label{sec:methodology}

For our performance evaluation, we used multiple, state-of-the-art  transformers that are on multiple natural language processing tasks.
For classification, we used a neural network containing the embedding layer and one hidden layer with a dropout layer between them.

\subsection{Transformers}

For our performance evaluation, we used multiple transformers that employed transfer learning to achieve state-of-the-art performance on multiple natural language processing tasks. 

\subsubsection{BERT}\label{sec:bert}
Bidirectional encoder representations from transformers (BERT),~\cite{Devlin2019}, is a deep bidirectional transformer architecture that employs transfer learning to generalize language understanding and can be fine-tuned on the dataset at hand.
Classic language models treat textual data as unidirectional or bidirectional sequences of words.
Instead, BERT uses the deep bidirectional transformer architecture to learn contextual relations between the words (or sub-words).
BERT reads the entire sequence of words at once using the transformer encoder. 
Thus, the model manages to learn the context of a word based on all of the surrounding words.
The experimental results show that the language models built with BERT manage to better determine the language context than the models that treat textual data as sequences of words. 

\subsubsection{DistilBERT}\label{sec:distilbert}
DistilBERT (Distilled BERT)~\cite{Sanh2019} is a method used to pre-train a smaller general-purpose language representation model by reducing the size of BERT.
As BERT, DistilBERT can be fine-tuned with good performances to solve different natural language processing tasks.
DistilBERT uses a triple loss combining language modeling, distillation, and cosine-distance losses.
The experimental results prove that DistilBERT is a smaller, faster, and lighter model than BERT, which manages to retain the language understanding capabilities.

\subsubsection{RoBERTa}\label{sec:roberta}
A robustly optimized BERT pre-training approach (RoBERTa)~\cite{Liu2019} is a robustly optimized method that improves BERT's language masking strategy by modifying key hyperparameters:
\begin{itemize}
    \item[(\textit{1})] It removes the next-sentence pre-training objective; and
    \item[(\textit{2})] It increases both the mini-batches and learning rates.
\end{itemize}

Thus, these modifications improve RoBERTa's masked language modeling objective, leading to better downstream task performance.

\subsubsection{DistilRoBERTa}\label{sec:distilroberta}
DistilRoBERTa (Distilled RoBERTa)~\cite{Sanh2019,Sajjad2020},
similar to DistilBERT, is a method used to pre-train a smaller general-purpose language representation model by reducing the size of RoBERTa.
It uses the same methodology employed to distill BERT on the RoBERTa~transformer.

\subsubsection{ALBERT}\label{sec:albert}
A lite BERT (ALBERT)~\cite{Lan2020} is another optimized method that improves on BERT. 
It reduces the number of parameters in order to lower memory consumption and increase training speed.
The authors of this method show through experimental validation that ALBERT leads to models that scale much better compared to BERT.
Furthermore, ALBERT uses a self-supervised loss that focuses on modeling inter-sentence coherence, which consistently improves downstream tasks with multi-sentence inputs.

\subsubsection{DeBERTa}\label{sec:deberta}
Decoding-enhanced BERT with disentangled attention (DeBERTa)~\cite{He2021} is an improvement of BERT and RoBERTa models.
It uses the disentangled attention mechanism, which employs two vectors to represent each word.
The first vector encodes the word's content, while the second its position.
Thus, the attention weights among words are computed using disentangled matrices on their contents and relative positions.
Moreover, DeBERTa uses an enhanced mask decoder to replace the output softmax layer to predict the masked tokens for model pre-training. 

\subsubsection{XLNet}\label{sec:xlnet}
XLNet~\cite{Yang2019} is an extension of BERT that integrates ideas from the transformer-XL model~\cite{Dai2019}.
It uses an autoregressive method to learn bidirectional contexts by maximizing the expected likelihood over all permutations of the input sequence factorization order.
Thus, XLNet overcomes the limitations of BERT through the autoregressive~formulation.

\subsubsection{ELECTRA}\label{sec:electra}
Efficiently learning an encoder that classifies token replacements accurately (ELECTRA)~\cite{Clark2020} is a pre-training approach used to build two transformer models: the generator and the discriminator.
The generator replaces tokens in a sequence and trains as a masked language model.
The discriminator predicts which tokens are replaced by the generator in the sequence.
We used ELECTRA's discriminator for classification.

\subsubsection{XLM}\label{sec:xlm}
The cross-lingual language model (XLM)~\cite{Conneau2019} is a transformer architecture that employs, during pre-training, either an unsupervised modeling technique that only relies on monolingual data, i.e., casual language modeling (CLM) or masked language modeling (MLM), or a supervised modeling technique that leverages parallel data with a new cross-lingual language model objective, i.e., MLM used in combination with translation language modeling (TLM).
CLM models the probability of a word given the previous words in a sentence, while MLM uses the masked language modeling of BERT.
TLM is an extension of MLM that concatenates parallel sentences instead of considering monolingual text streams.
This model manages to obtain better results than the baseline on multiple tasks, such as cross-lingual classification, unsupervised machine translation, supervised machine translation, low-resource language modeling, and unsupervised cross-lingual word embeddings.

\subsubsection{XLM-RoBERTa}\label{sec:xlmroberta}
XLM-RoBERTa~\cite{Conneau2020} is a transformer-based pre-trained multilingual masked language model.
The novelty of this model is that it does not require language tensors to understand which language is used.
Thus, the model is able to determine the correct language just from the input~IDs.

\subsubsection{BART}\label{sec:bart}
The bidirectional and autoregressive transformer (BART)~\cite{Lewis2020} uses a standard transformer-based neural machine translation architecture, which can be seen as generalizing BERT.
Thus, BART uses a standard sequence-to-sequence for machine translation architecture with a bidirectional encoder, similar to BERT, and a left-to-right decoder, similar to GPT.
BART is trained by corrupting text with an arbitrary noising function, and~learning a model to reconstruct the original text.
During the pre-training task, the model randomly shuffles the order of the original sentences. 
Moreover, the spans of text are replaced with a single mask token using a novel in-filling scheme.
BART is particularly effective when fine-tuned for text generation but it also works well for comprehension tasks~\cite{Lewis2020}.

\subsection{Classification Network}

The network for classification extracts the last hidden layer of the transformer, except for Electra, which uses the discriminator hidden states.
All networks use a final dense layer containing perceptrons for classification.
Between the transformer model and the classification layer, we used a dropout layer to prevent overfitting. 
The number of perceptrons in the classification layer is equal to the number of classes.
This final hidden layer, added after the transformer model and dropout layer, outputs the label for the multi-class classification.
DistilBERT and DistilRoBERTa also use a pre-classification, a ReLU, and a dropout layer between the output of the transformer model and the classification layer.
The pre-classification layer size is equal to the dimension of the encoder layers and the pooler layer.
The pre-classification layer is used to stabilize the output of the transformer and archive an improved classification performance~\cite{Seide2011}.
For XLM, XLM-RoBERTa, XLNet models, we used a sequence summary layer, which takes a sequence's hidden states and computes a single vector summary, followed by a dense layer for classification.

\section{MisRoBÆRTa: Misinformation RoBERTa-BART Ensemble Detection~Model}

\begin{figure}[!htbp]
\centering
\includegraphics[width=1\columnwidth]{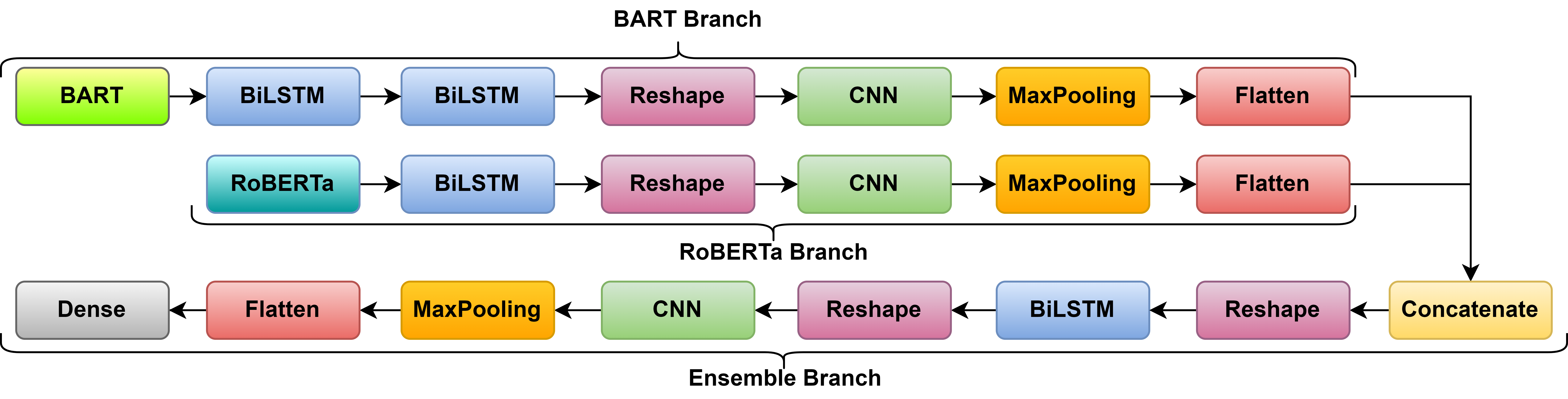}
\caption{MisRoBÆRTa Architecture.}
\label{fig:architecture}
\end{figure}

In this section, we propose a new deep learning architecture that develops hierarchical representations of learning using multiple bidirectional long short-term memory (BiLSTM) and convolutional layers (CNN).
Figure~\ref{fig:architecture} presents the computational workflow of our novel architecture MisRoBÆRTa: misinformation using RoBERTa-Bart sentence embeddings.
In our model, we have two inputs: 
\begin{itemize}
    \item[(\textit{1})] The sentence embeddings constructed with RoBERTa base of size 768; and
    \item[(\textit{2})] The sentence embeddings extracted with BART large of size 1024.
\end{itemize}

We chose these two transformers for MisRoBÆRTa because they had the best overall results among the pretrained transformers (Section~\ref{sec:results}) when the classification was done using the simple neural network model (Section~\ref{sec:methodology}).
Each embedding was then the input to a new block (branch) formed using the following layers: BiLSTM, reshape, CNN, max pooling, and~flatten.
During the ablation testing presented in Subsection~\ref{ss:ablation}
we obtained better results when employing two BiLSTM layers for the BART branch.
The output of the flatten layers was then concatenated, reshaped, and sent to another block (ensemble branch) containing the same layers that had, as input, the RoBERTa embeddings (RoBERTa branch).
Finally, there was a dense layer for the output of the classification.

\subsection{MisRoBÆRTa Components}

In this subsection, we present MisRoBÆRTa's main deep learning layers: dense, BiLSTM, and CNN.
We will not present again RoBERTa and BART, as both transformers are presented in Sections~\ref{sec:roberta} and~\ref{sec:bart}, respectively.

\subsubsection{Long Short-Term Memory Networks}

The long short-term memory network (LSTM)~\cite{Hochreiter1997} was developed to mitigate the vanishing or the exploding gradient problems that the classical recurrent neural network encountered.
By introducing gates within the cell, the LSTM manages to better capture dependencies.
The LSTM model uses the \textit{ {input gate}} $i_{t}$ for deciding what relevant information can be added from the current step $t$, using the previous hidden state $h_{t}$ and the input $x_{t}$, \textit{forget gate} $f_t$ for deciding which information is used and which is ignored in the current step $t$, and \textit{output gate} $o_{t}$ determines the value of the next hidden state $h_{t}$.
Equation~\eqref{eq:lstm_gates} presents of each gate (i.e., $f_{t}$, $i_{t}$, $o_{t}$) the current hidden state ($h_{t_1}$) and current memory state ($c_{t}$), where:
\begin{itemize}
    \item[(1)] $x_{t}$ is the input at time step $t$;
    \item[(2)] $h_{t}$ is the output, or next hidden state;
    \item[(3)] $h_{t-1}$ is the previous hidden state;
    \item[(4)] $\tilde{c}_{t}$ is the cell input activation vector;
    \item[(5)] $c_{t}$ is the current memory state;
    \item[(6)] $c_{t-1}$ is the previous memory state;
    \item[(7)] $W_{f}$, $W_{c}$, $W_{i}$, and $W_{o}$ are the weights for each gate's current input;
    \item[(8)] $U_{f}$, $U_{c}$, $U_{i}$, and $U_{o}$ are the weights for each gate's previous hidden state;
    \item[(9)] $b_{f}$, $b_{c}$, $b_{i}$, and $b_{o}$ are the bias vectors;
    \item[(10)] $\sigma_{s}$ is the sigmoid activation function;
    \item[(11)] $\sigma_{h}$ is the hyperbolic tangent activation function;
    \item[(12)] $\odot$ operator is the Hadamard product, i.e., the element-wise multiplication function.
\end{itemize}
\begin{equation}
\begin{split}
    i_{t} & = \sigma_{s}( W_{i}x_{t} + U_{i}h_{t-1} + b_{i} ) \\
    f_{t} & = \sigma_{s}( W_{f}x_{t} + U_{f}h_{t-1} + b_{f} ) \\
    o_{t} & = \sigma_{s}( W_{o}x_{t} + U_{o}h_{t-1} + b_{t} ) \\
    \tilde{c}_{t} & = \sigma_{h}( W_{c}x_{t} + U_{c}h_{t-1} + b_{c} ) \\
    c_{t} & = f_{t} \odot c_{t-1} + i_{t} * \tilde{c}_{t} \\
    h_{t} & = o_{t} \odot \tanh(c_{t})
    \label{eq:lstm_gates}
\end{split}
\end{equation}

The LSTM processes sequences and it is able to capture \textit{past} information.
To also consider \textit{future} information when building the model, the \textit{{bidirectional LSTM}} 
(BiLSTM) uses two LSTM with two hidden states $\overrightarrow{h_t}$ and $\overleftarrow{h_t}$ (Equation~\eqref{eq:bilstm}).
The first hidden state $\overrightarrow{h_t}$ processes the input in a forward manner using the past information provided by the forward LSTM $\overrightarrow{LSTM_F}$).
The second hidden state $\overleftarrow{h_t}$ processes the input in a backwards manner using the future information provided by the backward LSTM ($\overleftarrow{LSTM_B}$).
\begin{equation}\label{eq:bilstm}
\begin{split}
    \overrightarrow{h_t} & = \overrightarrow{LSTM_F}(x_t) \\
    \overleftarrow{h_t}  & = \overleftarrow{LSTM_B}(x_t)
\end{split}
\end{equation}

Finally, the hidden states $\overrightarrow{h_t}$ and $\overleftarrow{h_t}$ are concatenated at every time-step to enable encoding the information from past and future contexts into one hidden state $h'_t$ (Equation~\eqref{eq:concat}).

\begin{equation}\label{eq:concat}
    h'_t = [\overrightarrow{h_t} || \overleftarrow{h_t}]
\end{equation}

\subsubsection{Convolutional Neural Networks}

Unlike recurrent neural networks, which are used to detect patterns in sequences, convolutional neural networks (CNNs) were developed to detect patterns in hyperspaces.
In text classification, one-dimensional CNNs are used to apply a window of convolution that can extract n-grams.
Equation~\eqref{eq:cnn} presents a convolution unit $r_i$, where $x_i$ is a vector of filtered words,  ${\bm W}$ is a weight matrix, $b$ is the bias, and $\sigma$ is a non-linear activation function.

\begin{equation}\label{eq:cnn}
    r_{i} = \sigma(\bm{W} \cdot x_{i} + b)
\end{equation}

\subsubsection{Max Pooling}

The max pooling mechanism captures the most useful features produced by the previous layer.
The mechanism involves sliding a two-dimensional filter over each channel of the feature map, summarizing the features lying within the region covered by the filter. 
This layer applies an affine transformation to the feature vector to produce logits $d$.
Equation~(\ref{eq:maxpooling}) presents the logits $d$ outputted when using a feature map with the dimensions $n_h \times n_w \times n_c$, where $n_h$, $n_w$, $n_c$ are the height, the width, and the number of channels of the feature map,  respectively, $f$ is the size of the filter, and $s$ is the stride length.

\begin{equation}\label{eq:maxpooling}
    d = \frac{(n_h - f + 1)}{s} \times \frac{(n_w - f + 1)}{s} \times n_c
\end{equation} 

\subsubsection{Dense Layer}

The dense layer uses perceptron units to classify the output from the previous layer.
This layer received the information from the previous layer.
Equation~(\ref{eq:perceptron}) presents the representation of the perceptron predicted label $\hat{y}$, where $x_{i}$ is the input,  $\bm{w}$ is a weight vector, $b$ represents the bias, and $\delta(\cdot)$ is the activation function.
The activation function takes the total input and produces an output for the unit given some threshold.
In our model, we use the softmax activation function $\delta(x)=\frac{e^x}{\sum_{i=1}^{m}e^{x_i}}$ with $m$ the dimension of vector $x$.

\begin{equation}\label{eq:perceptron}
    \hat{y_{i}}=\delta(\bm{w} \cdot x_{i} + b)
\end{equation}

A loss function (Equation~\eqref{eq:loss_fn}) is used at each iteration to adjust the weights to align the predicted label $\hat{y}$ to the true label $y$.

\begin{equation}
    \label{eq:loss_fn}
    L(\hat{y}, y) = -(y\log\hat{y} + (1 - y) \log(1 - \hat{y}))
\end{equation}

\subsection{MisRoBÆRTa Description}

The BiLSTM layers contain 256 units each, 128 units for each long short-term memory (LSTM), and a dropout rate of 0.2.
We used the hyperbolic tangent as the activation function and the sigmoid as the recurrent activation function.
The CNN layers contain 64 filters, a kernel size of 128, and ReLU as the activation function.
After each CNN, we used a max pooling layer to down-sample the output and reduce the number.
Flatten was used to transform the two-dimensional output of the max pooling layer into a one-dimensional tensor.
The reshape layers were used to reshape the tensors to be accepted as input for the next layers.
Between the BiLSTM and the CNN, we needed a two-dimensional tensor, while between the concatenate layer and the BiLSTM, we needed a three-dimensional tensor.
The dense layer was a fully connected layer of perceptrons used for classification.
All parameters were determined using hyperparameter tuning.

\subsection{MisRoBÆRTa Architectural Choices}

Unidirectional LSTM preserves input information that has passed through its hidden state.
Thus, it only uses passed information to make assessments and correct predictions.
A BiLSTM instead preserves input information in two ways.
As LSTM, the BiLSTM will preserve the information that passes through the hidden layer, i.e., past to future.
Unlike LSTM, the BiLSTM also preserves information using the output of the hidden layer, and running backwards the preserved information, i.e., future to past.
Thus, BiLSTM combines two hidden states to preserve information from both past and future.
BiLSTM is better suited than LSTM to understand and preserved context. 
This is also supported by the ablation testing we present in Subsection~\ref{ss:ablation}.
Furthermore, we observed experimentally that we obtained better results when using two BiLSTM layers instead of one for the BART branch.

After the last BiLSTM layer for each branch, we employed a CNN layer.
CNNs have been used successfully for different text classifications and natural language tasks~\cite{Guo2019}.
When using a CNN, the output of each convolution detects patterns of multiple sizes by varying
the kernel shapes.
These outputs are then concatenated to permit the detection of multiple size patterns of adjacent words.
These patterns could be seen as multi-word expressions.
Therefore, CNNs can accurately identify and group multi-word expressions in the sentence regardless of their position.
These multi-word expressions better represent the context and the representative features of the input text.

\section{Experimental Results}\label{sec:results}

\subsection{Dataset}

The dataset used for experiments was the open-source \href{https://github.com/several27/FakeNewsCorpus}{\textsc{ {FakeNewsCorpus}}}  {(Accessed on the 6\textsuperscript{th} August 2021)} dataset available on GitHub~\cite{FakeNewsCorpus}.
It is comprised of more than 9.4 million labeled English news articles that contain both textual content (e.g., title,  content) and metadata (e.g., publisher, authors).
The original dataset is not manually filtered; therefore, some of the labels might not be correct.
However, the authors argue that because the corpus is intended to be used in training machine learning algorithms, this shortcoming should not pose a practical issue.
Moreover, this issue should help the models better generalize and remove overfitting.
For our experiments, we used only the news content and its labels.
Moreover, we made sure that the news sources contained articles that belonged to the type given by the class and we manually annotated each news article's content and compared it with its original label.
The annotation of the textual content was done by computer science students. 
We used two annotators for each article, introducing a third if there was no consensus between the first two.
The annotation task had two main limitations: author bias~\cite{Geva2019,Kuwatly2020} and incomplete annotations~\cite{Jie2019}.
Annotator bias refers to the habits and backgrounds of annotators that label the data and how these can impact the quality of the annotation and, as a result, the model.
To mitigate this, we presented the annotators with 10 classes from which to pick only one.
Thus the annotator, cannot bring any of his habits into the annotation process.
Incomplete annotations refer to the fact that the annotators may not agree among them.
In this situation, we added a third annotator who could choose only one of the two labels selected by the previous two annotators.
In this way, we mitigated disagreements among the annotators.
At the end of this task, we did not find
discrepancies between the original labels and the ones determined by the annotators.

Furthermore, we checked that the selected article URLs were pointing to the correct article by matching the titles and authors.
We did not apply preprocessing. 
The final dataset contained 10 classes (currently the 11{th} class contains 0 articles), with each class represented by 10,000 documents selected at random (Table~\ref{tab:classes}).
We chose this dataset because it is large enough to use with transformers.
We also presented the ablation tests for the \href{https://www.kaggle.com/c/fake-news/data}{ {Kaggle Fake News dataset}} (Accessed on the 9\textsuperscript{th} of October 2021) in Subsection~\ref{ss:ablation} used in~\cite{Kaliyar2020,Kaliyar2021}.

\begin{table}[!htbp]
\caption{News article classes.}
\centering
\label{tab:classes}
\begin{tabular}{ll}
\hline
\textbf{Class}        & \textbf{Description}                      \\ \hline
Fake News             & Fabricated or distorted information       \\ 
Satire                & Humorous or ironic information            \\ 
Extreme Bias          & Propaganda articles                       \\ 
Conspiracy Theory     & Promote conspiracy theories               \\ 
Junk Science          & Scientifically dubious claims             \\ 
Hate News             & Promote discrimination                    \\ 
Clickbait             & Credible content, but~misleading headline \\ 
Proceed With Caution  & May be reliable or not                    \\ 
Political             & Promote political orientations            \\ 
Credible              & Reliable information                      \\ \hline
\end{tabular}
\end{table}

Table~\ref{tab:desc} presents the dataset's description and token statistics (\#Tokens).
We observe that the distribution of tokens among the different classes is almost equal.
The satire class has the shortest textual content on average, while Hate News has the largest mean number of tokens.
We also note that the entire dataset contains 1,337,708 unique tokens and a total of 59,726,182 tokens.
By analyzing the dataset, we can affirm that there is no real bias added to the classification problem by the length of the documents.

\begin{table}[!htbp]
\centering
\caption{Dataset description and~statistics.}
\label{tab:desc}
\begin{tabular}{lrrrr|rr}

\hline
\multirow{2}{*}{\textbf{Class}\vspace{-5pt}} &  
\multicolumn{4}{c}{\textbf{Number of Tokens per Document}} & 
\multicolumn{2}{|c}{\textbf{Number of Tokens per Class}} \\ \cline{2-7}
      & \multicolumn{1}{c}{\textbf{Mean}}    & \multicolumn{1}{c}{\textbf{Min}} & \multicolumn{1}{c}{\textbf{Max }}    &  \multicolumn{1}{c}{\textbf{StdDev}}       & \multicolumn{1}{|c}{\textbf{Unique}} & \multicolumn{1}{c}{\textbf{All}} \\ \hline
\multicolumn{1}{l}{Fake News}             & 660.86  &  7  & 18,179  &    899.56  & 318,808 & 6,808,099 \\ 
\multicolumn{1}{l}{Satire}                & 245.19  & 11  &  4942  &    232.01  & 153,509 & 2,502,701 \\ 
\multicolumn{1}{l}{Extreme Bias}          & 539.71  &  5  & 17,402  & 1192.27  & 308,827 & 5,549,908 \\ 
\multicolumn{1}{l}{Conspiracy Theory}     & 800.17  &  7  & 17,448  &    947.65  & 315,146 & 8,182,202 \\ 
\multicolumn{1}{l}{Junk Science}          & 511.05  &  7  & 12,716  &    732.69  & 253,004 & 5,208,868 \\ 
\multicolumn{1}{l}{Hate News}             & 930.92  &  5  & 17,871  & 2064.67  & 348,295 & 9,516,495 \\ 
\multicolumn{1}{l}{Clickbait}             & 361.20  &  7  &  5544  &    361.62  & 193,844 & 3,702,766 \\ 
\multicolumn{1}{l}{Unreliable Sources}    & 505.63  &  7  & 17,232  &    982.76  & 236,495 & 5,149,370 \\ 
\multicolumn{1}{l}{Political Bias}        & 547.17  &  9  & 15,221  &    795.81  & 258,269 & 5,581,168 \\ 
\multicolumn{1}{l}{Credible}              & 737.98  &  6  & 15,049  &    818.02  & 257,475 & 7,524,605 \\  \hline
\multicolumn{1}{l}{\textbf{ {Entire dataset statistics}}} 
& 583.99  &  5  & 18,179  & 1037.13  & 1337,708 & 59,726,182 \\ \hline
\end{tabular}
\end{table}

To better understand the token distribution among the different classes, we analyzed documents using unigrams.
Table~\ref{tab:unigrams} presents the top-10 unigrams extracted using the \href{https://www.nltk.org/}{ {NLTK}}  {(Accessed on the 10\textsuperscript{th} of October 2021)}
python package~\cite{Bird2009}. 
We extracted the top-10 unigrams for the entire corpus as well as the top-10 unigrams for each class.
We used the cosine similarity, based on BERT and FastText, to measure the similarity between the unigrams extracted for the entire corpus and each class.
In our implementation, we used \href{https://maartengr.github.io/PolyFuzz/}{\textsc{ {PolyFuzz}}}  {(Accessed on the 14\textsuperscript{th} of October  2021)}~\cite{PolyFuzz}.
We observed that the smallest similarity was obtained when comparing the unigrams for the entire corpus with the ones extracted for the satire class, i.e.,~0.63 for BERT similarity and 0.44 for FastText similarity.
The rest of results are above 0.76 for BERT similarity and 0.67 for FastText similarity.
These results show that the terms appearing in each class are well represented within the entire corpus, although for the satire class, are underrepresented.
Based on the unigram analysis, we can conclude that there is no real bias added to classes by specific terms, as the distribution of the most frequent terms was similar. 

To better understand the context, we used a non-negative matrix factorization (NMF)~\cite{Arora2012} to extract the top-one topic for the entire dataset and each class.
To extract the topics, we cleaned the corpus removing stopwords and punctuation, then we vectorized the documents using the TFIDF. 
For our implementation, we used the TFIDF and NMF models from the \href{https://scikit-learn.org/stable/}{\textsc{ {scikit-learn}}}  {(Accessed on the 6\textsuperscript{th} of October 2021)}~\cite{Pedregosa2011} library.
This analysis was performed in order to determine if the underline hidden context present for each class would influence the behavior of the transformers during fine-tuning.
We observed that the best similarity was obtained by the topic extracted for the satire class (Table~\ref{tab:tm}).
The other classes had BERT-based similarities over 0.63.
These results show that the majority of articles contain context that is present among all classes, which reduces the bias added by the frequent topic and hidden context while preserving some features that can differentiate between the articles.

\begin{table}[!htbp]
\caption{Top-10 unigram similarity between the entire dataset and each~class.}
\label{tab:unigrams}
\resizebox{1\columnwidth}{!}{%
		\begin{tabular}{llrr}
\hline
\textbf{Class}     & \textbf{Top-10 Unigrams}   & \multirow{2}{*}{\begin{tabular}[c]{@{}r@{}} \textbf{Bert} \\ \textbf{Similarity} \end{tabular}\vspace{-5pt}} & \multirow{2}{*}{\begin{tabular}[c]{@{}r@{}} \textbf{FastText} \\ \textbf{Similarity} \end{tabular}\vspace{-5pt}} \\ \cline{1-2}
\textbf{Entire Dataset}     & \textbf{People Time Government American World System Year America State Public}     &
 &  \\ \hline
Fake News          & people time government world year story market American day God            & 0.89                                                                                                     & 0.84                                                                                                         \\
Satire             & order close continue user policy agree send deny click advertising         & 0.63                                                                                                     & 0.44                                                                                                         \\
Extreme Bias       & talk people American government time Russia America state war country      & 0.87                                                                                                     & 0.84                                                                                                         \\
Conspiracy         & people government time American America world report year country state    & 0.94                                                                                                     & 0.91                                                                                                         \\
Junk Science       & health people food free time world found body cancer study                 & 0.76                                                                                                     & 0.67                                                                                                         \\
Hate News          & people  black white snip percent American whites time country race         & 0.79                                                                                                     & 0.71                                                                                                         \\
Clickbait          & people Donald time state Clinton government America told country American  & 0.87                                                                                                     & 0.84                                                                                                         \\
Unreliable Sources & system Tor operating computer submission stick people public time order    & 0.79                                                                                                     & 0.71                                                                                                         \\
Political Bias     & people government time American percent year state president tax political & 0.89                                                                                                     & 0.83                                                                                                         \\
Credible           & people God Christian government American time world war told Iraq          & 0.85                                                                                                     & 0.81                                                                                                         \\ \hline
		\end{tabular}}
\end{table}

\begin{table}[!htbp]
\caption{Top-1 Topic similarity between the entire dataset and each~class.}
\label{tab:tm}
\resizebox{1\columnwidth}{!}{%
		\begin{tabular}{llrr}
\hline
\textbf{Class}     & \textbf{Top-1 Topic}   & \multirow{2}{*}{\begin{tabular}[c]{@{}r@{}}\textbf{Bert}\\ \textbf{Similarity}\end{tabular}\vspace{-5pt}} & \multirow{2}{*}{\begin{tabular}[c]{@{}r@{}}\textbf{FastText}\\ \textbf{Similarity}\end{tabular}\vspace{-5pt}} \\ \cline{1-2}
\textbf{Entire Dataset}     & \textbf{Banner Preference Navigation Consent Technical Advertising Profile Element User Click}   &                                                                                     &                                                                                         \\ \hline
Fake News          & not people Trump year day government time state world no                                & 0.63                                                                                & 0.41                                                                                    \\ 
Satire             & navigation banner advertising consent technical preference element click profile access & 0.97                                                                                & 0.95                                                                                    \\ 
Extreme Bias       & talk hide link page category template user supply file previous                         & 0.71                                                                                & 0.51                                                                                    \\ 
Conspiracy         & not report government people president year world state no time                         & 0.65                                                                                & 0.41                                                                                    \\ 
Junk Science       & food free health reference offer documentary herb nutrition list program                & 0.68                                                                                & 0.42                                                                                    \\ 
Hate News          & not black white snip people year percent school no student                              & 0.64                                                                                & 0.41                                                                                    \\ 
Clickbait          & Trump not president people twitter video state woman year no                            & 0.64                                                                                & 0.41                                                                                    \\ 
Unreliable Sources & Tor system tail browser submission computer stick communication GNU bundle              & 0.70                                                                                & 0.45                                                                                    \\ 
Political Bias     & Trump not president year state people house government no white                         & 0.64                                                                                & 0.41                                                                                    \\ 
Credible           & not church people Trump God president no war Bush state                                 & 0.63                                                                                & 0.39                                                                                    \\ 	\hline
		\end{tabular}}
\end{table}

Based on the analysis we performed using token statistics, unigram, and~topic modeling, we can conclude the following for the dataset.
Firstly, the length of the documents and the number of tokens for each class did not add any bias for the models to discriminate based on these two dimensions.
Secondly, the unique tokens per class versus the unique tokens for the entire dataset shows that some tokens were representative for some classes.
Thirdly, the most frequent unigrams did not influence the models, as the similarity between the unigrams extracted for each class and for the entire corpus was high.
Finally, the global context for each class had a moderate similarity with the context for the entire corpus. 
These results show that, among the different classes, there were some clear discriminative features.

\subsection{Fine-Tuning}
We used the \href{https://simpletransformers.ai}{\textsc{ {SimpleTransformers}}}  {(Accessed on the 12\textsuperscript{th} of October 2021)}~\cite{SimpleTransformers} library and the pre-trained transformers from \href{https://huggingface.co/}{\textsc{ {HuggingFace}}}  {(Accessed on the 9\textsuperscript{th} of October 2021)}~\cite{Wolf2020}.
We extended \textsc{SimpleTransformers} with our own implementation for a BART-based multi-class classifier.
Table~\ref{tab:trans} presents the transformers employed for the experiments.
The ALBERT v2 models did not have dropout.
We used the ELECTRA's discriminator for~classification.

\begin{table}[!htbp]
\centering
\caption{Transformer details.}
\label{tab:trans}
	\resizebox{1\columnwidth}{!}{%
		\begin{tabular}{llrrrr}
\hline
\textbf{Model} & \textbf{Pretrained Model Name}                 & \textbf{Encoder} \textbf{Layers} & \textbf{Hidden} \textbf{State} & \textbf{Attention} \textbf{Heads} & \textbf{Parameters} \\ \hline
BERT base           & bert-base-cased                     & 12                      & 768                  & 12                       & 110~M                \\ 
BERT large          & bert-large-cased                    & 24                      & 1024               & 16                       & 335~M                \\ 
DistilBERT          & distilbert-base-cased               & 6                       & 768                  & 12                       &  65~M                \\ 
RoBERTa base        & roberta-base                        & 12                      & 768                  & 12                       & 125~M                \\ 
RoBERTa large       & roberta-large                       & 24                      & 1024               & 16                       & 355~M                \\ 
DistilRoBERTa       & distilroberta-base                  &  6                      & 768                  & 12                       &  82~M                \\ 
XLNet base          & xlnet-base-cased                    & 12                      & 768                  & 12                       & 110~M                \\ 
XLNet large         & xlnet-large-cased                   & 24                      & 1024               & 16                       & 340~M                \\ 
ALBERT base v1      & albert-base-v1                      & 12                      & 768                  & 12                       &  11~M                \\ 
ALBERT base v2      & albert-base-v2                      & 12                      & 768                  & 12                       &  11~M                \\ 
ALBERT xxlarge v1   & albert-xxlarge-v1                   & 12                      & 4096               & 64                       & 223~M                \\ 
ALBERT xxlarge v2   & albert-xxlarge-v2                   & 12                      & 4096               & 64                       & 223~M                \\ 
DeBERTa base        & microsoft/deberta-base              & 12                      & 768                  & 12                       & 140~M                \\ 
DeBERTa large       & microsoft/deberta-large             & 24                      & 1024               & 16                       & 400~M                \\ 
ELECTRA base        & google/electra-base-discriminator   & 12                      & 768                  & 12                       & 110~M                \\ 
ELECTRA large       & google/electra-large-discriminator  & 24                      & 1024               & 16                       & 335~M                \\ 
XLM                 & xlm-mlm-100-1280                    & 16                      & 1280               & 16                       & 550~M                \\ 
XLM-RoBERTa base  & xlm-roberta-base                      & 12                      & 768                  & 8                        & 270~M                \\ 
XLM-RoBERTa large & lm-roberta-large                      & 24                      & 1024               & 16                       & 550~M                \\ 
BART base           & facebook/bart-base                  & 12                      & 768                  & 16                       & 139~M                \\ 
BART large          & facebook/bart-large                 & 12                      & 1024               & 16                       & 406~M                \\ \hline
		\end{tabular}
  }
\end{table}

The employed transformers were fine-tuned by training them on our dataset for 10~epochs with an early stopping mechanism that evaluated the loss over 5 epochs.
We split the dataset using a 64\%--16\%--20\% training--validation--testing ratio with random seeding.
For each split, we maintained the stratification of the labels.
We applied this method for 10~rounds of~tests.

\subsection{Classification Results}

We used an NVIDIA\textsuperscript{\textregistered} DGX Station\textsuperscript{\texttrademark} containing 4 NVIDIA\textsuperscript{\textregistered} V100 Tensor Core GPUs for our experiments.
Table~\ref{tab:results} presents the classification results.
We observed that all transformer models had an accuracy of over 85\%, with the lowest accuracy obtained by ALBERT base v2.
For comparison reasons, we also tested the state-of-the-art model FakeBERT~\cite{Kaliyar2021}. 
As the authors of this model did not provide their implementation, we implemented FakeBERT using BERT base together with the \href{https://www.tensorflow.org/}{\textsc{ {TensorFlow}}}  {(Accessed on 8\textsuperscript{th} October 2021)}~\cite{tensorflow2015-whitepaper} and \href{https://keras.io/}{\textsc{ {Keras}}}  {(Accessed on the 8\textsuperscript{th} of October 2021)}~\cite{chollet2015keras} libraries. 
The code for the MisRoBÆRTa, transformer implementation, and FakeBERT are publicly available on GitHub at \url{https://github.com/cipriantruica/MisRoBAERTa_Transformers-vs-misinformation}.

The BERT, RoBERTa, DeBERTa, XLNet, ELECTRA, and~XLM-RoBERTa base models obtained better results than their counterpart large models.
For these models, we can conclude that a smaller number of encoder layers with smaller sizes better generalized and managed to correctly extract  the hidden context within the texts.

The reverse happened for ALBERT and BART, where the large models outperformed the base models. 
Although the difference in performance between BART base and BART large was small ($\sim$0.60\%) for ALBERT, the difference was between $\sim$1.30\% and $\sim$2.3\%.
For BART, we can conclude that the autoregressive method, the sequence-to-sequence architecture, and the left--right decoder played an important factor and improved performance as the model's number of layers and size increased.
For ALBERT, the self-supervised loss that focused on modeling inter-sentence coherence improved the model's performance as the network size increased. 
Furthermore, the use of dropout improved ALBERT performance significantly, as ALBERT v2 versions had the overall worst performance.

We observe that DeBERTa obtained slightly better results than BERT, regardless if it was the base or the large model.  
DeBERTa base obtained an accuracy slightly worst than the RoBERTa base, while DeBERTa large outperformed RoBERTa large.
XLM obtained the second worst accuracy among the tested~transformers.

\begin{table}[!htbp]
\centering
\caption{Misinformation classification results (note: italic text marks similar results while bold text shows the overall best result).}
\label{tab:results}
\resizebox{1\columnwidth}{!}{%
		\begin{tabular}{lcccccc}
\hline
\textbf{Model}                   & \textbf{Accuracy}            & \textbf{Micro Precision}    & \textbf{Macro Precision}  & \textbf{Micro Recall} & \textbf{Macro Recall} & \textbf{Execution Time (Hours)}  \\ \hline
MisRoBÆRTa                       & \textbf{92.50 $\pm$ 0.26}    & 92.50 $\pm$ 0.26            & 92.69 $\pm$ 0.21          & 92.50 $\pm$ 0.26      & 92.50 $\pm$ 0.26      &  1.83 $\pm$ 0.01           \\ 
BERT base                        & 90.03 $\pm$ 0.19             & 90.03 $\pm$ 0.19            & 90.05 $\pm$ 0.21          & 90.03 $\pm$ 0.19      & 90.03 $\pm$ 0.19      &  5.12 $\pm$ 0.22           \\ 
BERT large                       & 89.12 $\pm$ 0.14             & 89.12 $\pm$ 0.14            & 89.11 $\pm$ 0.15          & 89.12 $\pm$ 0.14      & 89.12 $\pm$ 0.14      &  8.97 $\pm$ 0.05           \\ 
DistilBERT                       & 89.36 $\pm$ 0.15             & 89.36 $\pm$ 0.15            & 89.40 $\pm$ 0.16          & 89.36 $\pm$ 0.15      & 89.36 $\pm$ 0.15      &  2.30 $\pm$ 0.01           \\ 
RoBERTa base                     & \textit{91.36 $\pm$ 0.15}    & 91.36 $\pm$ 0.15            & 91.39 $\pm$ 0.16          & 91.36 $\pm$ 0.15      & 91.36 $\pm$ 0.15      &  3.95 $\pm$ 0.15           \\ 
RoBERTa large                    & 88.60 $\pm$ 0.23             & 88.61 $\pm$ 0.22            & 88.85 $\pm$ 0.23          & 88.59 $\pm$ 0.24      & 88.62 $\pm$ 0.21      &  6.55 $\pm$ 0.32           \\ 
DistilRoBERTa                    & \textit{91.32 $\pm$ 0.10}    & 91.32 $\pm$ 0.10	          & 91.34 $\pm$ 0.08	      & 91.32 $\pm$ 0.10	  & 91.32 $\pm$ 0.10      &  1.93 $\pm$ 0.01           \\ 
ALBERT base v1                   & 86.92 $\pm$ 0.17             & 86.92 $\pm$ 0.17            & 86.95 $\pm$ 0.18          & 86.92 $\pm$ 0.17      & 86.92 $\pm$ 0.17      &  2.16 $\pm$ 0.19           \\ 
ALBERT base v2                   & 85.05 $\pm$ 0.20             & 85.05 $\pm$ 0.20            & 85.08 $\pm$ 0.17          & 85.05 $\pm$ 0.20      & 85.05 $\pm$ 0.20      &  2.76 $\pm$ 0.17           \\ 
ALBERT xxlarge v1                & 89.20 $\pm$ 0.04             & 89.20 $\pm$ 0.04            & 89.36 $\pm$ 0.02          & 89.20 $\pm$ 0.04      & 89.20 $\pm$ 0.04      &  9.98 $\pm$ 0.09           \\ 
ALBERT xxlarge v2                & 86.51 $\pm$ 2.23             & 86.51 $\pm$ 2.23            & 86.66 $\pm$ 2.21          & 86.51 $\pm$ 2.23      & 86.51 $\pm$ 2.23      &       11.14 $\pm$ 0.04\,\,\,           \\ 
DeBERTa base                     & 90.56 $\pm$ 0.08             & 90.56 $\pm$ 0.08            & 90.56 $\pm$ 0.10          & 90.56 $\pm$ 0.08      & 90.56 $\pm$ 0.08      &  6.90 $\pm$ 0.35           \\ 
DeBERTa large                    & 89.93 $\pm$ 0.27             & 89.93 $\pm$ 0.27            & 89.96 $\pm$ 0.29          & 89.93 $\pm$ 0.27      & 89.93 $\pm$ 0.27      &       11.06 $\pm$ 1.10\,\,\,           \\ 
XLNet base                       & 89.94 $\pm$ 0.09             & 89.94 $\pm$ 0.09            & 89.94 $\pm$ 0.09          & 89.94 $\pm$ 0.09      & 89.94 $\pm$ 0.09      &  5.50 $\pm$ 0.06           \\ 
XLNet large                      & 88.05 $\pm$ 0.39             & 88.04 $\pm$ 0.38            & 88.38 $\pm$ 0.39          & 88.03 $\pm$ 0.37      & 88.07 $\pm$ 0.40      &       10.44 $\pm$ 1.54\,\,\,           \\ 
ELECTRA base                     & 87.10 $\pm$ 0.24             & 87.10 $\pm$ 0.24            & 87.12 $\pm$ 0.22          & 87.10 $\pm$ 0.24      & 87.10 $\pm$ 0.24      &  4.49 $\pm$ 0.34           \\ 
ELECTRA large                    & 86.92 $\pm$ 0.09             & 86.92 $\pm$ 0.09            & 86.92 $\pm$ 0.08          & 86.92 $\pm$ 0.09      & 86.92 $\pm$ 0.09      &  8.74 $\pm$ 0.43           \\ 
XLM	                             & 85.82 $\pm$ 0.19             & 85.82 $\pm$ 0.19            & 85.90 $\pm$ 0.23          & 85.82 $\pm$ 0.19      & 85.82 $\pm$ 0.19      &  5.16 $\pm$ 0.01           \\ 
XLM-RoBERTa base                 & 89.78 $\pm$ 0.17             & 89.78 $\pm$ 0.17            & 89.77 $\pm$ 0.17          & 89.78 $\pm$ 0.17      & 89.78 $\pm$ 0.17      &  6.97 $\pm$ 0.38           \\ 
XLM-RoBERTa large                & 87.50 $\pm$ 0.61             & 87.50 $\pm$ 0.61            & 87.58 $\pm$ 0.58          & 87.50 $\pm$ 0.61      & 87.50 $\pm$ 0.61      &  9.46 $\pm$ 2.16           \\ 
BART base                        & \textit{91.35 $\pm$ 0.04}    & 91.35 $\pm$ 0.04            & 91.35 $\pm$ 0.04          & 91.35 $\pm$ 0.04      & 91.35 $\pm$ 0.04      &  4.17 $\pm$ 0.16           \\ 
BART large                       & \textit{91.94 $\pm$ 0.15}    & 91.94 $\pm$ 0.15            & 91.97 $\pm$ 0.16          & 91.94 $\pm$ 0.15      & 91.94 $\pm$ 0.15      &  6.79 $\pm$ 0.29           \\ 
FakeBERT~\cite{Kaliyar2021}      & 70.18 $\pm$ 0.01             & 70.18 $\pm$ 0.01            & 70.21 $\pm$ 0.03          & 70.18 $\pm$ 0.01      & 70.18 $\pm$ 0.01      &  2.59 $\pm$ 0.01           \\ \hline
		\end{tabular}}
\end{table}

Both DistilBERT and DistilRoBERTa were outperformed by BERT, respectively, RoBERTa base models. 
The difference in accuracy between the base and distilled models was very small: $\sim$0.04\% between DistilBERT and BERT base and $\sim$0.70\% between DistilRoBERTa and RoBERTa base.
Although DistilBERT outperformed BERT large with a small increase in accuracy ($\sim$0.24\%), DistilRoBERTa outperformed RoBERTa large with a large increase in accuracy ($\sim$2.72\%).
We can conclude that, for the use case of misinformation classification, distillation is not a good approach if the computational and memory resources needed for the base approaches are available.

The micro and macro recall metrics are the most relevant for misinformation classification, as they show the correctly classified news articles relative to all the news articles, regardless of the predicted label.
The micro and macro precisions manage to show the percentage of the articles, which are relevant.
All  transformers obtained results over 85\% for these four metrics, meaning that they managed to generalize well for the task of misinformation~detection.

The overall best performing model was the novel MisRoBÆRTa with an accuracy of 92.50\%, while the overall best performing transformer model was BART large, with an accuracy of 91.94\%.
The next three transformers that obtained accuracies over 91\% were RoBERTa base (91.36\%), BART base (91.35\%), and~DistilRoBERTa 91.32\%.

We observed that the state-of-the-art model FakeBERT had very low performance on our dataset. 
Although the original paper shows that FakeBERT obtained an accuracy of 98.9\% on a binary dataset from Kaggle, in our multi-class dataset, it obtained only $\sim$70\%. 
These results might be a consequence of the following: 
\begin{itemize}
    \item[(\textit{1})] FakeBERT is developed for binary classification, while we used it for multi-class classification;
    \item[(\textit{2})] To change FakeBERT from a binary classifier to a multi-class classifier, we modified the number of units of its final DENSE layer from 2 to the number of classes, i.e.,~10;
    \item[(\textit{3})] We only trained FakeBERT for 10 epochs, to be consistent in our performance experiments (note: this is also the number of epochs used in the original paper).
\end{itemize}

MisRoBÆRTa had the overall best classification results with an accuracy of 92.50\%.
This accuracy was a direct result of the way context was preserved through the use of:
\begin{itemize}
    \item[(\textit{1})] BART large and RoBERTa base;
    \item[(\textit{2})] BiLSTM layers that combined two hidden states to preserve information from both past and future;
    \item[(\textit{3})] CNNs to generate multi-word expressions, which better represent the feature space. 
\end{itemize}

\begin{figure}[!htbp]
\centering
\includegraphics[width=1\columnwidth]{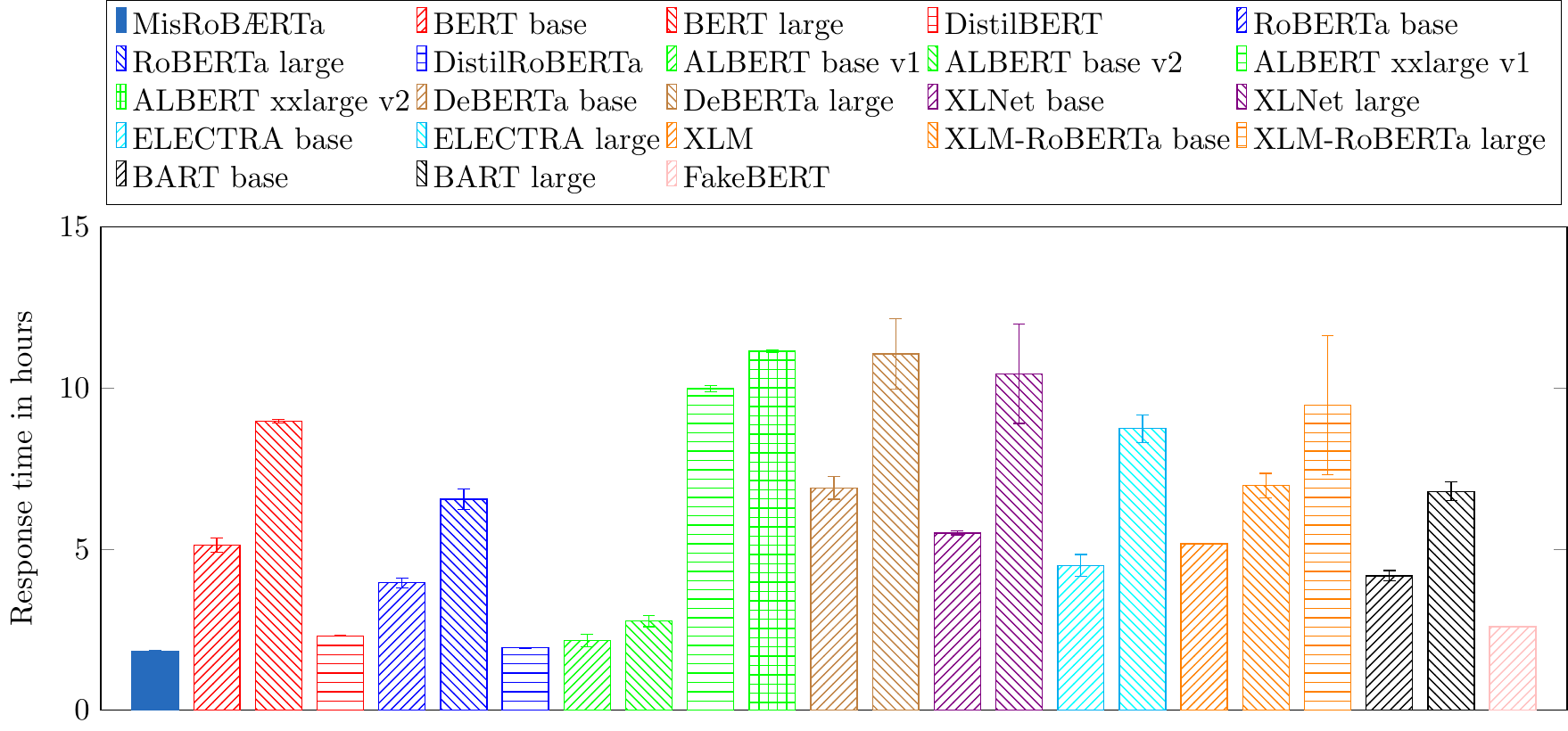}
\caption{Performance time for the tested architectures.}
\label{fig:performance}
\end{figure}

\subsection{Runtime Evaluation}

To evaluate the runtime, we ran each transformer for 10 epochs without early stopping.
As already mentioned, we used an NVIDIA\textsuperscript{\textregistered} DGX Station\textsuperscript{\texttrademark} containing 4 NVIDIA\textsuperscript{\textregistered} V100 Tensor Core GPUs for our evaluation.
Figure~\ref{fig:performance} presents the mean execution time and the standard deviation for 10 rounds of tests.
As expected, the large models are much slower than their base counterparts. 
Moreover, we note that the distilled versions DistilRoBERTa, ALBERT base, and~DistilBERT outperformed both BERT and RoBERTa.
MisRoBÆRTa had the fastest runtime among the tested models managing to converge in 1.83h (this included the maximum time taken to create the sentence embeddings in parallel using the two transformers: 1.53h for BART and 0.35h for RoBERTa).

For the base models, the fastest transformer model was DistilRoBERTa, which required, on average, 1.93~h for fine-tuning and training, while the slowest was XLM-RoBERTa, which finished training and fine-tuning in a little less than 7h on average.
For the large models, RoBERTa large was the fastest, requiring on average 6.55h for fine-tuning and training, while the slowest was ALBERT large v2, which finished in 11.14h on average.
FakeBERT converged in 2.59h on average.
This result was directly impacted by the CNN and max pooling layers it employed to minimize the input space.
XLM runtime performance at 5.16~h on average was between BERT base and XLNet base.
In conclusion, the fastest transformer model that also achieved a high accuracy was DistilRoBERTa.

\subsection{Ablation Testing}\label{ss:ablation}

In Tables~\ref{tab:ablation2} and~\ref{tab:ablation}, we present ablation and sensitivity testing for MisRoBÆRTa using the publicly available \href{https://www.kaggle.com/c/fake-news/data}{ {Kaggle Fake News Dataset}} and the FakeNewsCorpus dataset presented in Section~\ref{sec:results}. The tests include (1) testing with LSTM and BiLSTM units, (2) variations on the number of layers for each branch, e.g., 1, 2, and 3 layers, and (3) variation in the number of units for each LSTM layer, e.g., 32, 64, 128.
The results are aggregated after 10 runs, for each execution, the dataset was split at random without replacement, but keeping the class ratio into 64\% training, 16\% validation, and 20\% testing.
We present only the network's training time, the sentence embedding creation time was not taken into account. 
We should note that the results we obtained during ablation and sensitivity testing on the publicly available \href{https://www.kaggle.com/c/fake-news/data}{ {Kaggle Fake News dataset}} are similar to the current state-of-the-art results obtained on this corpus by~\cite{Kaliyar2021,Kaliyar2020}.

\begin{table}[!htbp]
\caption{MisRoBÆRTa ablation and sensitivity testing using the \href{https://www.kaggle.com/c/fake-news/data}{ {Kaggle Fake News Dataset.}} 
(Note: bold text marks the overall best result).} 
\label{tab:ablation2}
\resizebox{\textwidth}{!}{%
		\begin{tabular}{ccccccccccccccc}
\hline
\multicolumn{3}{c}{\textbf{32 Units/Layer}} & \multicolumn{6}{c}{\textbf{MisRoBÆRTa with LSTM Cells}} & \multicolumn{6}{c}{\textbf{MisRoBÆRTa with BiLSTM Cells}} \\ 
\hline
\textbf{\begin{tabular}[c]{@{}c@{}}BART\\ Branch\end{tabular}} & \textbf{\begin{tabular}[c]{@{}c@{}}RoBERTa\\ Branch\end{tabular}} & \textbf{\begin{tabular}[c]{@{}c@{}}Ensemble\\ Branch\end{tabular}} & \textbf{Accuracy} & \textbf{\begin{tabular}[c]{@{}c@{}}Precision\\ Micro\end{tabular}} & \textbf{\begin{tabular}[c]{@{}c@{}}Precision\\ Macro\end{tabular}} & \textbf{\begin{tabular}[c]{@{}c@{}}Recall\\ Micro\end{tabular}} & \textbf{\begin{tabular}[c]{@{}c@{}}Recall\\ Macro\end{tabular}} & \textbf{\begin{tabular}[c]{@{}c@{}}Execution\\ Time (Hours)\end{tabular}} & \textbf{Accuracy} & \textbf{\begin{tabular}[c]{@{}c@{}}Precision\\ Micro\end{tabular}} & \textbf{\begin{tabular}[c]{@{}c@{}}Precision\\ Macro\end{tabular}} & \textbf{\begin{tabular}[c]{@{}c@{}}Recall\\ Micro\end{tabular}} & \textbf{\begin{tabular}[c]{@{}c@{}}Recall\\ Macro\end{tabular}} & \textbf{\begin{tabular}[c]{@{}c@{}}Execution\\ Time (Hours)\end{tabular}} \\ 
\hline
1 { $\times$ }[Bi]LSTM 
& 1 $\times$ [Bi]LSTM & 1 $\times$ [Bi]LSTM & 97.10 $\pm$ 0.43 & 97.10 $\pm$ 0.43 & 97.16 $\pm$ 0.42 & 97.10 $\pm$ 0.43 & 97.07 $\pm$ 0.43 & 0.01 $\pm$ 0.00 & 97.35 $\pm$ 0.37 & 97.35 $\pm$ 0.37 & 97.40 $\pm$ 0.36 & 97.35 $\pm$ 0.37 & 97.33 $\pm$ 0.37 & 0.01 $\pm$ 0.00 \\ 
1 $\times$ [Bi]LSTM & 1 $\times$ [Bi]LSTM & 2 $\times$ [Bi]LSTM & 97.14 $\pm$ 0.35 & 97.14 $\pm$ 0.35 & 97.17 $\pm$ 0.35 & 97.14 $\pm$ 0.35 & 97.12 $\pm$ 0.35 & 0.02 $\pm$ 0.00 & 97.30 $\pm$ 0.42 & 97.30 $\pm$ 0.42 & 97.33 $\pm$ 0.38 & 97.30 $\pm$ 0.42 & 97.29 $\pm$ 0.45 & 0.03 $\pm$ 0.00 \\ 
1 $\times$ [Bi]LSTM & 1 $\times$ [Bi]LSTM & 3 $\times$ [Bi]LSTM & 97.10 $\pm$ 0.57 & 97.10 $\pm$ 0.57 & 97.13 $\pm$ 0.56 & 97.10 $\pm$ 0.57 & 97.08 $\pm$ 0.57 & 0.03 $\pm$ 0.00 & 97.32 $\pm$ 0.37 & 97.32 $\pm$ 0.37 & 97.34 $\pm$ 0.35 & 97.32 $\pm$ 0.37 & 97.34 $\pm$ 0.36 & 0.05 $\pm$ 0.00 \\ 
1 $\times$ [Bi]LSTM & 2 $\times$ [Bi]LSTM & 1 $\times$ [Bi]LSTM & 97.06 $\pm$ 0.28 & 97.06 $\pm$ 0.28 & 97.09 $\pm$ 0.28 & 97.06 $\pm$ 0.28 & 97.05 $\pm$ 0.27 & 0.04 $\pm$ 0.00 & 97.02 $\pm$ 0.48 & 97.02 $\pm$ 0.48 & 97.10 $\pm$ 0.44 & 97.02 $\pm$ 0.48 & 96.99 $\pm$ 0.49 & 0.06 $\pm$ 0.00 \\ 
1 $\times$ [Bi]LSTM & 2 $\times$ [Bi]LSTM & 2 $\times$ [Bi]LSTM & 97.15 $\pm$ 0.35 & 97.15 $\pm$ 0.35 & 97.16 $\pm$ 0.35 & 97.15 $\pm$ 0.35 & 97.15 $\pm$ 0.35 & 0.05 $\pm$ 0.00 & 97.43 $\pm$ 0.28 & 97.43 $\pm$ 0.28 & 97.44 $\pm$ 0.26 & 97.43 $\pm$ 0.28 & 97.43 $\pm$ 0.30 & 0.08 $\pm$ 0.00 \\ 
1 $\times$ [Bi]LSTM & 2 $\times$ [Bi]LSTM & 3 $\times$ [Bi]LSTM & 97.16 $\pm$ 0.37 & 97.16 $\pm$ 0.37 & 97.18 $\pm$ 0.37 & 97.16 $\pm$ 0.37 & 97.17 $\pm$ 0.35 & 0.07 $\pm$ 0.00 & 97.36 $\pm$ 0.32 & 97.36 $\pm$ 0.32 & 97.38 $\pm$ 0.30 & 97.36 $\pm$ 0.32 & 97.38 $\pm$ 0.33 & 0.11 $\pm$ 0.00 \\ 
1 $\times$ [Bi]LSTM & 3 $\times$ [Bi]LSTM & 1 $\times$ [Bi]LSTM & 97.09 $\pm$ 0.46 & 97.09 $\pm$ 0.46 & 97.10 $\pm$ 0.45 & 97.09 $\pm$ 0.46 & 97.11 $\pm$ 0.44 & 0.08 $\pm$ 0.00 & 97.33 $\pm$ 0.32 & 97.33 $\pm$ 0.32 & 97.35 $\pm$ 0.30 & 97.33 $\pm$ 0.32 & 97.32 $\pm$ 0.34 & 0.13 $\pm$ 0.00 \\ 
1 $\times$ [Bi]LSTM & 3 $\times$ [Bi]LSTM & 2 $\times$ [Bi]LSTM & 97.07 $\pm$ 0.37 & 97.07 $\pm$ 0.37 & 97.08 $\pm$ 0.38 & 97.07 $\pm$ 0.37 & 97.08 $\pm$ 0.36 & 0.09 $\pm$ 0.00 & 97.14 $\pm$ 0.41 & 97.14 $\pm$ 0.41 & 97.16 $\pm$ 0.39 & 97.14 $\pm$ 0.41 & 97.17 $\pm$ 0.39 & 0.15 $\pm$ 0.00 \\ 
1 $\times$ [Bi]LSTM & 3 $\times$ [Bi]LSTM & 3 $\times$ [Bi]LSTM & 97.30 $\pm$ 0.37 & 97.30 $\pm$ 0.37 & 97.35 $\pm$ 0.38 & 97.30 $\pm$ 0.37 & 97.27 $\pm$ 0.36 & 0.11 $\pm$ 0.00 & 97.24 $\pm$ 0.53 & 97.24 $\pm$ 0.53 & 97.31 $\pm$ 0.48 & 97.24 $\pm$ 0.53 & 97.22 $\pm$ 0.55 & 0.18 $\pm$ 0.01 \\ 
2 $\times$ [Bi]LSTM & 1 $\times$ [Bi]LSTM & 1 $\times$ [Bi]LSTM & \textbf{97.56 $\pm$ 0.29} & 97.56 $\pm$ 0.29 & 97.57 $\pm$ 0.27 & 97.56 $\pm$ 0.29 & 97.55 $\pm$ 0.30 & 0.12 $\pm$ 0.00 & \textbf{97.57 $\pm$ 0.29} & 97.57 $\pm$ 0.29 & 97.58 $\pm$ 0.28 & 97.57 $\pm$ 0.29 & 97.57 $\pm$ 0.31 & 0.19 $\pm$ 0.01 \\ 
2 $\times$ [Bi]LSTM & 1 $\times$ [Bi]LSTM & 2 $\times$ [Bi]LSTM & 97.34 $\pm$ 0.36 & 97.34 $\pm$ 0.36 & 97.38 $\pm$ 0.35 & 97.34 $\pm$ 0.36 & 97.32 $\pm$ 0.36 & 0.13 $\pm$ 0.00 & 97.44 $\pm$ 0.38 & 97.44 $\pm$ 0.38 & 97.44 $\pm$ 0.38 & 97.44 $\pm$ 0.38 & 97.44 $\pm$ 0.37 & 0.21 $\pm$ 0.01 \\ 
2 $\times$ [Bi]LSTM & 1 $\times$ [Bi]LSTM & 3 $\times$ [Bi]LSTM & 97.20 $\pm$ 0.38 & 97.20 $\pm$ 0.38 & 97.24 $\pm$ 0.36 & 97.20 $\pm$ 0.38 & 97.18 $\pm$ 0.39 & 0.15 $\pm$ 0.00 & 97.31 $\pm$ 0.44 & 97.31 $\pm$ 0.44 & 97.34 $\pm$ 0.42 & 97.31 $\pm$ 0.44 & 97.30 $\pm$ 0.46 & 0.23 $\pm$ 0.01 \\ 
2 $\times$ [Bi]LSTM & 2 $\times$ [Bi]LSTM & 1 $\times$ [Bi]LSTM & 97.31 $\pm$ 0.47 & 97.31 $\pm$ 0.47 & 97.38 $\pm$ 0.41 & 97.31 $\pm$ 0.47 & 97.28 $\pm$ 0.50 & 0.16 $\pm$ 0.00 & 97.41 $\pm$ 0.37 & 97.41 $\pm$ 0.37 & 97.42 $\pm$ 0.38 & 97.41 $\pm$ 0.37 & 97.41 $\pm$ 0.37 & 0.25 $\pm$ 0.01 \\ 
2 $\times$ [Bi]LSTM & 2 $\times$ [Bi]LSTM & 2 $\times$ [Bi]LSTM & 97.44 $\pm$ 0.26 & 97.44 $\pm$ 0.26 & 97.45 $\pm$ 0.26 & 97.44 $\pm$ 0.26 & 97.42 $\pm$ 0.27 & 0.18 $\pm$ 0.00 & 97.26 $\pm$ 0.37 & 97.26 $\pm$ 0.37 & 97.30 $\pm$ 0.36 & 97.26 $\pm$ 0.37 & 97.25 $\pm$ 0.36 & 0.28 $\pm$ 0.01 \\ 
2 $\times$ [Bi]LSTM & 2 $\times$ [Bi]LSTM & 3 $\times$ [Bi]LSTM & 97.30 $\pm$ 0.33 & 97.30 $\pm$ 0.33 & 97.34 $\pm$ 0.25 & 97.30 $\pm$ 0.33 & 97.29 $\pm$ 0.35 & 0.19 $\pm$ 0.00 & 97.40 $\pm$ 0.44 & 97.40 $\pm$ 0.44 & 97.44 $\pm$ 0.42 & 97.40 $\pm$ 0.44 & 97.37 $\pm$ 0.44 & 0.30 $\pm$ 0.01 \\ 
2 $\times$ [Bi]LSTM & 3 $\times$ [Bi]LSTM & 1 $\times$ [Bi]LSTM & 97.37 $\pm$ 0.41 & 97.37 $\pm$ 0.41 & 97.40 $\pm$ 0.41 & 97.37 $\pm$ 0.41 & 97.36 $\pm$ 0.41 & 0.21 $\pm$ 0.00 & 97.40 $\pm$ 0.37 & 97.40 $\pm$ 0.37 & 97.42 $\pm$ 0.36 & 97.40 $\pm$ 0.37 & 97.39 $\pm$ 0.37 & 0.33 $\pm$ 0.01 \\ 
2 $\times$ [Bi]LSTM & 3 $\times$ [Bi]LSTM & 2 $\times$ [Bi]LSTM & 97.39 $\pm$ 0.36 & 97.39 $\pm$ 0.36 & 97.39 $\pm$ 0.37 & 97.39 $\pm$ 0.36 & 97.39 $\pm$ 0.36 & 0.22 $\pm$ 0.00 & 97.46 $\pm$ 0.22 & 97.46 $\pm$ 0.22 & 97.47 $\pm$ 0.20 & 97.46 $\pm$ 0.22 & 97.47 $\pm$ 0.20 & 0.35 $\pm$ 0.01 \\ 
2 $\times$ [Bi]LSTM & 3 $\times$ [Bi]LSTM & 3 $\times$ [Bi]LSTM & 97.09 $\pm$ 0.35 & 97.09 $\pm$ 0.35 & 97.14 $\pm$ 0.34 & 97.09 $\pm$ 0.35 & 97.10 $\pm$ 0.33 & 0.24 $\pm$ 0.00 & 97.16 $\pm$ 0.37 & 97.16 $\pm$ 0.37 & 97.17 $\pm$ 0.38 & 97.16 $\pm$ 0.37 & 97.16 $\pm$ 0.36 & 0.38 $\pm$ 0.01 \\ 
3 $\times$ [Bi]LSTM & 1 $\times$ [Bi]LSTM & 1 $\times$ [Bi]LSTM & 97.43 $\pm$ 0.39 & 97.43 $\pm$ 0.39 & 97.46 $\pm$ 0.37 & 97.43 $\pm$ 0.39 & 97.41 $\pm$ 0.41 & 0.26 $\pm$ 0.00 & 97.45 $\pm$ 0.39 & 97.45 $\pm$ 0.39 & 97.48 $\pm$ 0.39 & 97.45 $\pm$ 0.39 & 97.44 $\pm$ 0.39 & 0.40 $\pm$ 0.01 \\ 
3 $\times$ [Bi]LSTM & 1 $\times$ [Bi]LSTM & 2 $\times$ [Bi]LSTM & 97.42 $\pm$ 0.33 & 97.42 $\pm$ 0.33 & 97.44 $\pm$ 0.34 & 97.42 $\pm$ 0.33 & 97.40 $\pm$ 0.32 & 0.27 $\pm$ 0.01 & 97.36 $\pm$ 0.35 & 97.36 $\pm$ 0.35 & 97.40 $\pm$ 0.35 & 97.36 $\pm$ 0.35 & 97.35 $\pm$ 0.35 & 0.43 $\pm$ 0.01 \\ 
3 $\times$ [Bi]LSTM & 1 $\times$ [Bi]LSTM & 3 $\times$ [Bi]LSTM & 97.19 $\pm$ 0.29 & 97.19 $\pm$ 0.29 & 97.30 $\pm$ 0.23 & 97.19 $\pm$ 0.29 & 97.15 $\pm$ 0.32 & 0.29 $\pm$ 0.00 & 97.41 $\pm$ 0.27 & 97.41 $\pm$ 0.27 & 97.42 $\pm$ 0.29 & 97.41 $\pm$ 0.27 & 97.40 $\pm$ 0.27 & 0.46 $\pm$ 0.01 \\ 
3 $\times$ [Bi]LSTM & 2 $\times$ [Bi]LSTM & 1 $\times$ [Bi]LSTM & 97.25 $\pm$ 0.44 & 97.25 $\pm$ 0.44 & 97.28 $\pm$ 0.41 & 97.25 $\pm$ 0.44 & 97.25 $\pm$ 0.42 & 0.31 $\pm$ 0.01 & 97.33 $\pm$ 0.47 & 97.33 $\pm$ 0.47 & 97.38 $\pm$ 0.44 & 97.33 $\pm$ 0.47 & 97.33 $\pm$ 0.43 & 0.48 $\pm$ 0.01 \\ 
3 $\times$ [Bi]LSTM & 2 $\times$ [Bi]LSTM & 2 $\times$ [Bi]LSTM & 97.31 $\pm$ 0.32 & 97.31 $\pm$ 0.32 & 97.36 $\pm$ 0.28 & 97.31 $\pm$ 0.32 & 97.29 $\pm$ 0.34 & 0.32 $\pm$ 0.01 & 97.30 $\pm$ 0.39 & 97.30 $\pm$ 0.39 & 97.34 $\pm$ 0.32 & 97.30 $\pm$ 0.39 & 97.29 $\pm$ 0.42 & 0.51 $\pm$ 0.01 \\ 
3 $\times$ [Bi]LSTM & 2 $\times$ [Bi]LSTM & 3 $\times$ [Bi]LSTM & 97.23 $\pm$ 0.41 & 97.23 $\pm$ 0.41 & 97.26 $\pm$ 0.40 & 97.23 $\pm$ 0.41 & 97.23 $\pm$ 0.38 & 0.34 $\pm$ 0.01 & 97.48 $\pm$ 0.42 & 97.48 $\pm$ 0.42 & 97.48 $\pm$ 0.42 & 97.48 $\pm$ 0.42 & 97.50 $\pm$ 0.41 & 0.54 $\pm$ 0.01 \\ 
3 $\times$ [Bi]LSTM & 3 $\times$ [Bi]LSTM & 1 $\times$ [Bi]LSTM & 97.41 $\pm$ 0.34 & 97.41 $\pm$ 0.34 & 97.43 $\pm$ 0.35 & 97.41 $\pm$ 0.34 & 97.39 $\pm$ 0.32 & 0.36 $\pm$ 0.01 & 97.31 $\pm$ 0.30 & 97.31 $\pm$ 0.30 & 97.35 $\pm$ 0.31 & 97.31 $\pm$ 0.30 & 97.30 $\pm$ 0.30 & 0.57 $\pm$ 0.01 \\ 
3 $\times$ [Bi]LSTM & 3 $\times$ [Bi]LSTM & 2 $\times$ [Bi]LSTM & 97.24 $\pm$ 0.29 & 97.24 $\pm$ 0.29 & 97.29 $\pm$ 0.31 & 97.24 $\pm$ 0.29 & 97.21 $\pm$ 0.28 & 0.38 $\pm$ 0.01 & 97.42 $\pm$ 0.45 & 97.42 $\pm$ 0.45 & 97.44 $\pm$ 0.43 & 97.42 $\pm$ 0.45 & 97.42 $\pm$ 0.46 & 0.60 $\pm$ 0.02 \\ 
3 $\times$ [Bi]LSTM & 3 $\times$ [Bi]LSTM & 3 $\times$ [Bi]LSTM & 97.15 $\pm$ 0.31 & 97.15 $\pm$ 0.31 & 97.17 $\pm$ 0.30 & 97.15 $\pm$ 0.31 & 97.13 $\pm$ 0.32 & 0.40 $\pm$ 0.01 & 97.30 $\pm$ 0.44 & 97.30 $\pm$ 0.44 & 97.32 $\pm$ 0.43 & 97.30 $\pm$ 0.44 & 97.31 $\pm$ 0.43 & 0.63 $\pm$ 0.02 \\

\hline
\multicolumn{3}{c}{\textbf{64 Units/Layer}} & \multicolumn{6}{c}{\textbf{MisRoBÆRTa with LSTM Cells}} & \multicolumn{6}{c}{\textbf{MisRoBÆRTa with BiLSTM Cells}} \\ 
\hline
\textbf{\begin{tabular}[c]{@{}c@{}}BART\\ Branch\end{tabular}} & \textbf{\begin{tabular}[c]{@{}c@{}}RoBERTa\\ Branch\end{tabular}} & \textbf{\begin{tabular}[c]{@{}c@{}}Ensemble\\ Branch\end{tabular}} & \textbf{Accuracy} & \textbf{\begin{tabular}[c]{@{}c@{}}Precision\\ Micro\end{tabular}} & \textbf{\begin{tabular}[c]{@{}c@{}}Precision\\ Macro\end{tabular}} & \textbf{\begin{tabular}[c]{@{}c@{}}Recall\\ Micro\end{tabular}} & \textbf{\begin{tabular}[c]{@{}c@{}}Recall\\ Macro\end{tabular}} & \textbf{\begin{tabular}[c]{@{}c@{}}Execution\\ Time (Hours)\end{tabular}} & \textbf{Accuracy} & \textbf{\begin{tabular}[c]{@{}c@{}}Precision\\ Micro\end{tabular}} & \textbf{\begin{tabular}[c]{@{}c@{}}Precision\\ Macro\end{tabular}} & \textbf{\begin{tabular}[c]{@{}c@{}}Recall\\ Micro\end{tabular}} & \textbf{\begin{tabular}[c]{@{}c@{}}Recall\\ Macro\end{tabular}} & \textbf{\begin{tabular}[c]{@{}c@{}}Execution\\ Time (Hours)\end{tabular}} \\ 
\hline
1 $\times$ [Bi]LSTM & 1 $\times$ [Bi]LSTM & 1 $\times$ [Bi]LSTM & 97.24 $\pm$ 0.78 & 97.24 $\pm$ 0.78 & 97.29 $\pm$ 0.72 & 97.24 $\pm$ 0.78 & 97.25 $\pm$ 0.77 & 0.01 $\pm$ 0.00 & 97.42 $\pm$ 0.47 & 97.42 $\pm$ 0.47 & 97.49 $\pm$ 0.40 & 97.42 $\pm$ 0.47 & 97.39 $\pm$ 0.48 & 0.01 $\pm$ 0.00 \\ 
1 $\times$ [Bi]LSTM & 1 $\times$ [Bi]LSTM & 2 $\times$ [Bi]LSTM & 97.36 $\pm$ 0.42 & 97.36 $\pm$ 0.42 & 97.41 $\pm$ 0.39 & 97.36 $\pm$ 0.42 & 97.34 $\pm$ 0.43 & 0.02 $\pm$ 0.00 & 97.29 $\pm$ 0.48 & 97.29 $\pm$ 0.48 & 97.35 $\pm$ 0.40 & 97.29 $\pm$ 0.48 & 97.29 $\pm$ 0.50 & 0.03 $\pm$ 0.00 \\ 
1 $\times$ [Bi]LSTM & 1 $\times$ [Bi]LSTM & 3 $\times$ [Bi]LSTM & 97.42 $\pm$ 0.47 & 97.42 $\pm$ 0.47 & 97.44 $\pm$ 0.41 & 97.42 $\pm$ 0.47 & 97.41 $\pm$ 0.48 & 0.03 $\pm$ 0.00 & 97.20 $\pm$ 0.50 & 97.20 $\pm$ 0.50 & 97.25 $\pm$ 0.46 & 97.20 $\pm$ 0.50 & 97.20 $\pm$ 0.48 & 0.04 $\pm$ 0.00 \\ 
1 $\times$ [Bi]LSTM & 2 $\times$ [Bi]LSTM & 1 $\times$ [Bi]LSTM & 97.48 $\pm$ 0.37 & 97.48 $\pm$ 0.37 & 97.48 $\pm$ 0.37 & 97.48 $\pm$ 0.37 & 97.48 $\pm$ 0.38 & 0.04 $\pm$ 0.00 & 97.26 $\pm$ 0.28 & 97.26 $\pm$ 0.28 & 97.36 $\pm$ 0.17 & 97.26 $\pm$ 0.28 & 97.24 $\pm$ 0.33 & 0.06 $\pm$ 0.00 \\ 
1 $\times$ [Bi]LSTM & 2 $\times$ [Bi]LSTM & 2 $\times$ [Bi]LSTM & 97.33 $\pm$ 0.39 & 97.33 $\pm$ 0.39 & 97.34 $\pm$ 0.39 & 97.33 $\pm$ 0.39 & 97.33 $\pm$ 0.38 & 0.05 $\pm$ 0.00 & 97.13 $\pm$ 0.39 & 97.13 $\pm$ 0.39 & 97.25 $\pm$ 0.30 & 97.13 $\pm$ 0.39 & 97.10 $\pm$ 0.43 & 0.08 $\pm$ 0.00 \\ 
1 $\times$ [Bi]LSTM & 2 $\times$ [Bi]LSTM & 3 $\times$ [Bi]LSTM & 97.36 $\pm$ 0.42 & 97.36 $\pm$ 0.42 & 97.38 $\pm$ 0.40 & 97.36 $\pm$ 0.42 & 97.35 $\pm$ 0.40 & 0.06 $\pm$ 0.00 & 97.35 $\pm$ 0.39 & 97.35 $\pm$ 0.39 & 97.40 $\pm$ 0.37 & 97.35 $\pm$ 0.39 & 97.34 $\pm$ 0.39 & 0.10 $\pm$ 0.00 \\ 
1 $\times$ [Bi]LSTM & 3 $\times$ [Bi]LSTM & 1 $\times$ [Bi]LSTM & 97.32 $\pm$ 0.46 & 97.32 $\pm$ 0.46 & 97.35 $\pm$ 0.45 & 97.32 $\pm$ 0.46 & 97.31 $\pm$ 0.46 & 0.07 $\pm$ 0.00 & 97.29 $\pm$ 0.77 & 97.29 $\pm$ 0.77 & 97.37 $\pm$ 0.67 & 97.29 $\pm$ 0.77 & 97.29 $\pm$ 0.73 & 0.12 $\pm$ 0.00 \\ 
1 $\times$ [Bi]LSTM & 3 $\times$ [Bi]LSTM & 2 $\times$ [Bi]LSTM & 97.53 $\pm$ 0.34 & 97.53 $\pm$ 0.34 & 97.56 $\pm$ 0.33 & 97.53 $\pm$ 0.34 & 97.50 $\pm$ 0.35 & 0.09 $\pm$ 0.00 & 97.20 $\pm$ 0.79 & 97.20 $\pm$ 0.79 & 97.30 $\pm$ 0.66 & 97.20 $\pm$ 0.79 & 97.22 $\pm$ 0.73 & 0.14 $\pm$ 0.00 \\ 
1 $\times$ [Bi]LSTM & 3 $\times$ [Bi]LSTM & 3 $\times$ [Bi]LSTM & 97.32 $\pm$ 0.40 & 97.32 $\pm$ 0.40 & 97.32 $\pm$ 0.39 & 97.32 $\pm$ 0.40 & 97.33 $\pm$ 0.38 & 0.10 $\pm$ 0.00 & 97.33 $\pm$ 0.50 & 97.33 $\pm$ 0.50 & 97.38 $\pm$ 0.45 & 97.33 $\pm$ 0.50 & 97.35 $\pm$ 0.47 & 0.16 $\pm$ 0.00 \\ 
2 $\times$ [Bi]LSTM & 1 $\times$ [Bi]LSTM & 1 $\times$ [Bi]LSTM & \textbf{97.84 $\pm$ 0.50} & 97.84 $\pm$ 0.50 & 97.85 $\pm$ 0.50 & 97.84 $\pm$ 0.50 & 97.84 $\pm$ 0.48 & 0.11 $\pm$ 0.00 & \textbf{97.85 $\pm$ 0.23} & 97.85 $\pm$ 0.23 & 97.88 $\pm$ 0.24 & 97.85 $\pm$ 0.23 & 97.83 $\pm$ 0.22 & 0.18 $\pm$ 0.00 \\ 
2 $\times$ [Bi]LSTM & 1 $\times$ [Bi]LSTM & 2 $\times$ [Bi]LSTM & 97.56 $\pm$ 0.38 & 97.56 $\pm$ 0.38 & 97.58 $\pm$ 0.40 & 97.56 $\pm$ 0.38 & 97.55 $\pm$ 0.37 & 0.12 $\pm$ 0.00 & 97.40 $\pm$ 0.38 & 97.40 $\pm$ 0.38 & 97.51 $\pm$ 0.34 & 97.40 $\pm$ 0.38 & 97.36 $\pm$ 0.38 & 0.20 $\pm$ 0.00 \\ 
2 $\times$ [Bi]LSTM & 1 $\times$ [Bi]LSTM & 3 $\times$ [Bi]LSTM & 97.56 $\pm$ 0.48 & 97.56 $\pm$ 0.48 & 97.60 $\pm$ 0.42 & 97.56 $\pm$ 0.48 & 97.56 $\pm$ 0.50 & 0.14 $\pm$ 0.00 & 97.39 $\pm$ 0.36 & 97.39 $\pm$ 0.36 & 97.45 $\pm$ 0.37 & 97.39 $\pm$ 0.36 & 97.37 $\pm$ 0.33 & 0.22 $\pm$ 0.00 \\ 
2 $\times$ [Bi]LSTM & 2 $\times$ [Bi]LSTM & 1 $\times$ [Bi]LSTM & 97.53 $\pm$ 0.48 & 97.53 $\pm$ 0.48 & 97.57 $\pm$ 0.48 & 97.53 $\pm$ 0.48 & 97.50 $\pm$ 0.47 & 0.15 $\pm$ 0.00 & 97.33 $\pm$ 0.46 & 97.33 $\pm$ 0.46 & 97.41 $\pm$ 0.35 & 97.33 $\pm$ 0.46 & 97.32 $\pm$ 0.50 & 0.24 $\pm$ 0.01 \\ 
2 $\times$ [Bi]LSTM & 2 $\times$ [Bi]LSTM & 2 $\times$ [Bi]LSTM & 97.56 $\pm$ 0.45 & 97.56 $\pm$ 0.45 & 97.61 $\pm$ 0.42 & 97.56 $\pm$ 0.45 & 97.55 $\pm$ 0.43 & 0.16 $\pm$ 0.00 & 97.38 $\pm$ 0.65 & 97.38 $\pm$ 0.65 & 97.42 $\pm$ 0.60 & 97.38 $\pm$ 0.65 & 97.38 $\pm$ 0.63 & 0.26 $\pm$ 0.01 \\ 
2 $\times$ [Bi]LSTM & 2 $\times$ [Bi]LSTM & 3 $\times$ [Bi]LSTM & 97.53 $\pm$ 0.38 & 97.53 $\pm$ 0.38 & 97.56 $\pm$ 0.36 & 97.53 $\pm$ 0.38 & 97.52 $\pm$ 0.39 & 0.18 $\pm$ 0.00 & 97.35 $\pm$ 0.41 & 97.35 $\pm$ 0.41 & 97.43 $\pm$ 0.34 & 97.35 $\pm$ 0.41 & 97.34 $\pm$ 0.41 & 0.28 $\pm$ 0.01 \\ 
2 $\times$ [Bi]LSTM & 3 $\times$ [Bi]LSTM & 1 $\times$ [Bi]LSTM & 97.52 $\pm$ 0.35 & 97.52 $\pm$ 0.35 & 97.55 $\pm$ 0.34 & 97.52 $\pm$ 0.35 & 97.52 $\pm$ 0.35 & 0.19 $\pm$ 0.00 & 97.47 $\pm$ 0.21 & 97.47 $\pm$ 0.21 & 97.52 $\pm$ 0.20 & 97.47 $\pm$ 0.21 & 97.44 $\pm$ 0.22 & 0.30 $\pm$ 0.01 \\ 
2 $\times$ [Bi]LSTM & 3 $\times$ [Bi]LSTM & 2 $\times$ [Bi]LSTM & 97.52 $\pm$ 0.37 & 97.52 $\pm$ 0.37 & 97.57 $\pm$ 0.34 & 97.52 $\pm$ 0.37 & 97.50 $\pm$ 0.39 & 0.21 $\pm$ 0.00 & 97.46 $\pm$ 0.34 & 97.46 $\pm$ 0.34 & 97.52 $\pm$ 0.29 & 97.46 $\pm$ 0.34 & 97.47 $\pm$ 0.31 & 0.33 $\pm$ 0.01 \\ 
2 $\times$ [Bi]LSTM & 3 $\times$ [Bi]LSTM & 3 $\times$ [Bi]LSTM & 97.63 $\pm$ 0.31 & 97.63 $\pm$ 0.31 & 97.65 $\pm$ 0.30 & 97.63 $\pm$ 0.31 & 97.60 $\pm$ 0.31 & 0.22 $\pm$ 0.01 & 97.46 $\pm$ 0.25 & 97.46 $\pm$ 0.25 & 97.49 $\pm$ 0.18 & 97.46 $\pm$ 0.25 & 97.46 $\pm$ 0.28 & 0.36 $\pm$ 0.01 \\ 
3 $\times$ [Bi]LSTM & 1 $\times$ [Bi]LSTM & 1 $\times$ [Bi]LSTM & 97.72 $\pm$ 0.25 & 97.72 $\pm$ 0.25 & 97.73 $\pm$ 0.26 & 97.72 $\pm$ 0.25 & 97.70 $\pm$ 0.25 & 0.24 $\pm$ 0.01 & 97.60 $\pm$ 0.37 & 97.60 $\pm$ 0.37 & 97.65 $\pm$ 0.40 & 97.60 $\pm$ 0.37 & 97.56 $\pm$ 0.37 & 0.38 $\pm$ 0.01 \\ 
3 $\times$ [Bi]LSTM & 1 $\times$ [Bi]LSTM & 2 $\times$ [Bi]LSTM & 97.67 $\pm$ 0.43 & 97.67 $\pm$ 0.43 & 97.72 $\pm$ 0.43 & 97.67 $\pm$ 0.43 & 97.65 $\pm$ 0.41 & 0.25 $\pm$ 0.01 & 97.66 $\pm$ 0.34 & 97.66 $\pm$ 0.34 & 97.66 $\pm$ 0.33 & 97.66 $\pm$ 0.34 & 97.66 $\pm$ 0.33 & 0.40 $\pm$ 0.01 \\ 
3 $\times$ [Bi]LSTM & 1 $\times$ [Bi]LSTM & 3 $\times$ [Bi]LSTM & 97.58 $\pm$ 0.43 & 97.58 $\pm$ 0.43 & 97.64 $\pm$ 0.35 & 97.58 $\pm$ 0.43 & 97.55 $\pm$ 0.47 & 0.27 $\pm$ 0.01 & 97.57 $\pm$ 0.31 & 97.57 $\pm$ 0.31 & 97.60 $\pm$ 0.32 & 97.57 $\pm$ 0.31 & 97.55 $\pm$ 0.31 & 0.42 $\pm$ 0.01 \\ 
3 $\times$ [Bi]LSTM & 2 $\times$ [Bi]LSTM & 1 $\times$ [Bi]LSTM & 97.59 $\pm$ 0.30 & 97.59 $\pm$ 0.30 & 97.62 $\pm$ 0.29 & 97.59 $\pm$ 0.30 & 97.58 $\pm$ 0.31 & 0.28 $\pm$ 0.01 & 97.52 $\pm$ 0.29 & 97.52 $\pm$ 0.29 & 97.54 $\pm$ 0.28 & 97.52 $\pm$ 0.29 & 97.50 $\pm$ 0.30 & 0.44 $\pm$ 0.01 \\ 
3 $\times$ [Bi]LSTM & 2 $\times$ [Bi]LSTM & 2 $\times$ [Bi]LSTM & 97.57 $\pm$ 0.37 & 97.57 $\pm$ 0.37 & 97.60 $\pm$ 0.38 & 97.57 $\pm$ 0.37 & 97.56 $\pm$ 0.36 & 0.30 $\pm$ 0.01 & 97.38 $\pm$ 0.47 & 97.38 $\pm$ 0.47 & 97.41 $\pm$ 0.43 & 97.38 $\pm$ 0.47 & 97.41 $\pm$ 0.46 & 0.47 $\pm$ 0.01 \\ 
3 $\times$ [Bi]LSTM & 2 $\times$ [Bi]LSTM & 3 $\times$ [Bi]LSTM & 97.63 $\pm$ 0.38 & 97.63 $\pm$ 0.38 & 97.65 $\pm$ 0.35 & 97.63 $\pm$ 0.38 & 97.63 $\pm$ 0.39 & 0.31 $\pm$ 0.01 & 97.56 $\pm$ 0.34 & 97.56 $\pm$ 0.34 & 97.57 $\pm$ 0.33 & 97.56 $\pm$ 0.34 & 97.57 $\pm$ 0.34 & 0.50 $\pm$ 0.01 \\ 
3 $\times$ [Bi]LSTM & 3 $\times$ [Bi]LSTM & 1 $\times$ [Bi]LSTM & 97.36 $\pm$ 0.48 & 97.36 $\pm$ 0.48 & 97.41 $\pm$ 0.47 & 97.36 $\pm$ 0.48 & 97.35 $\pm$ 0.46 & 0.33 $\pm$ 0.01 & 97.30 $\pm$ 0.48 & 97.30 $\pm$ 0.48 & 97.41 $\pm$ 0.42 & 97.30 $\pm$ 0.48 & 97.24 $\pm$ 0.50 & 0.52 $\pm$ 0.01 \\ 
3 $\times$ [Bi]LSTM & 3 $\times$ [Bi]LSTM & 2 $\times$ [Bi]LSTM & 97.62 $\pm$ 0.48 & 97.62 $\pm$ 0.48 & 97.65 $\pm$ 0.45 & 97.62 $\pm$ 0.48 & 97.61 $\pm$ 0.49 & 0.35 $\pm$ 0.01 & 97.52 $\pm$ 0.20 & 97.52 $\pm$ 0.20 & 97.54 $\pm$ 0.22 & 97.52 $\pm$ 0.20 & 97.50 $\pm$ 0.20 & 0.55 $\pm$ 0.01 \\ 
3 $\times$ [Bi]LSTM & 3 $\times$ [Bi]LSTM & 3 $\times$ [Bi]LSTM & 97.46 $\pm$ 0.35 & 97.46 $\pm$ 0.35 & 97.47 $\pm$ 0.35 & 97.46 $\pm$ 0.35 & 97.46 $\pm$ 0.36 & 0.37 $\pm$ 0.01 & 97.50 $\pm$ 0.39 & 97.50 $\pm$ 0.39 & 97.51 $\pm$ 0.38 & 97.50 $\pm$ 0.39 & 97.51 $\pm$ 0.37 & 0.58 $\pm$ 0.01 \\ 

\hline
\multicolumn{3}{c}{\textbf{128 Units/Layer}} & \multicolumn{6}{c}{\textbf{MisRoBÆRTa with LSTM Cells}} & \multicolumn{6}{c}{\textbf{MisRoBÆRTa with BiLSTM Cells}} \\ 
\hline
\textbf{\begin{tabular}[c]{@{}c@{}}BART\\ Branch\end{tabular}} & \textbf{\begin{tabular}[c]{@{}c@{}}RoBERTa\\ Branch\end{tabular}} & \textbf{\begin{tabular}[c]{@{}c@{}}Ensemble\\ Branch\end{tabular}} & \textbf{Accuracy} & \textbf{\begin{tabular}[c]{@{}c@{}}Precision\\ Micro\end{tabular}} & \textbf{\begin{tabular}[c]{@{}c@{}}Precision\\ Macro\end{tabular}} & \textbf{\begin{tabular}[c]{@{}c@{}}Recall\\ Micro\end{tabular}} & \textbf{\begin{tabular}[c]{@{}c@{}}Recall\\ Macro\end{tabular}} & \textbf{\begin{tabular}[c]{@{}c@{}}Execution\\ Time (Hours)\end{tabular}} & \textbf{Accuracy} & \textbf{\begin{tabular}[c]{@{}c@{}}Precision\\ Micro\end{tabular}} & \textbf{\begin{tabular}[c]{@{}c@{}}Precision\\ Macro\end{tabular}} & \textbf{\begin{tabular}[c]{@{}c@{}}Recall\\ Micro\end{tabular}} & \textbf{\begin{tabular}[c]{@{}c@{}}Recall\\ Macro\end{tabular}} & \textbf{\begin{tabular}[c]{@{}c@{}}Execution\\ Time (Hours)\end{tabular}} \\ 
\hline
1 $\times$ [Bi]LSTM & 1 $\times$ [Bi]LSTM & 1 $\times$ [Bi]LSTM & 97.17 $\pm$ 0.54 & 97.17 $\pm$ 0.54 & 97.22 $\pm$ 0.50 & 97.17 $\pm$ 0.54 & 97.17 $\pm$ 0.54 & 0.01 $\pm$ 0.00 & 97.23 $\pm$ 0.52 & 97.23 $\pm$ 0.52 & 97.26 $\pm$ 0.49 & 97.23 $\pm$ 0.52 & 97.25 $\pm$ 0.49 & 0.01 $\pm$ 0.00 \\ 
1 $\times$ [Bi]LSTM & 1 $\times$ [Bi]LSTM & 2 $\times$ [Bi]LSTM & 93.98 $\pm$ 0.97 & 93.98 $\pm$ 0.97 & 97.06 $\pm$ 0.84 & 93.98 $\pm$ 0.97 & 97.01 $\pm$ 0.92 & 0.02 $\pm$ 0.00 & 97.32 $\pm$ 0.27 & 97.32 $\pm$ 0.27 & 97.36 $\pm$ 0.24 & 97.32 $\pm$ 0.27 & 97.31 $\pm$ 0.29 & 0.03 $\pm$ 0.00 \\ 
1 $\times$ [Bi]LSTM & 1 $\times$ [Bi]LSTM & 3 $\times$ [Bi]LSTM & 97.04 $\pm$ 0.55 & 97.04 $\pm$ 0.55 & 97.10 $\pm$ 0.47 & 97.04 $\pm$ 0.55 & 97.04 $\pm$ 0.59 & 0.03 $\pm$ 0.00 & 97.19 $\pm$ 0.32 & 97.19 $\pm$ 0.32 & 97.21 $\pm$ 0.30 & 97.19 $\pm$ 0.32 & 97.20 $\pm$ 0.32 & 0.04 $\pm$ 0.00 \\ 
1 $\times$ [Bi]LSTM & 2 $\times$ [Bi]LSTM & 1 $\times$ [Bi]LSTM & 97.18 $\pm$ 0.61 & 97.18 $\pm$ 0.61 & 97.25 $\pm$ 0.60 & 97.18 $\pm$ 0.61 & 97.18 $\pm$ 0.58 & 0.03 $\pm$ 0.00 & 97.33 $\pm$ 0.42 & 97.33 $\pm$ 0.42 & 97.38 $\pm$ 0.44 & 97.33 $\pm$ 0.42 & 97.33 $\pm$ 0.42 & 0.06 $\pm$ 0.00 \\ 
1 $\times$ [Bi]LSTM & 2 $\times$ [Bi]LSTM & 2 $\times$ [Bi]LSTM & 97.13 $\pm$ 0.81 & 97.13 $\pm$ 0.81 & 97.19 $\pm$ 0.76 & 97.13 $\pm$ 0.81 & 97.14 $\pm$ 0.79 & 0.05 $\pm$ 0.00 & 97.37 $\pm$ 0.26 & 97.37 $\pm$ 0.26 & 97.37 $\pm$ 0.25 & 97.37 $\pm$ 0.26 & 97.38 $\pm$ 0.27 & 0.08 $\pm$ 0.00 \\ 
1 $\times$ [Bi]LSTM & 2 $\times$ [Bi]LSTM & 3 $\times$ [Bi]LSTM & 97.34 $\pm$ 0.48 & 97.34 $\pm$ 0.48 & 97.37 $\pm$ 0.47 & 97.34 $\pm$ 0.48 & 97.34 $\pm$ 0.47 & 0.06 $\pm$ 0.00 & 97.03 $\pm$ 0.86 & 97.03 $\pm$ 0.86 & 97.11 $\pm$ 0.72 & 97.03 $\pm$ 0.86 & 97.06 $\pm$ 0.80 & 0.10 $\pm$ 0.00 \\ 
1 $\times$ [Bi]LSTM & 3 $\times$ [Bi]LSTM & 1 $\times$ [Bi]LSTM & 97.38 $\pm$ 0.65 & 97.38 $\pm$ 0.65 & 97.40 $\pm$ 0.64 & 97.38 $\pm$ 0.65 & 97.37 $\pm$ 0.64 & 0.07 $\pm$ 0.00 & 97.04 $\pm$ 0.53 & 97.04 $\pm$ 0.53 & 97.14 $\pm$ 0.44 & 97.04 $\pm$ 0.53 & 97.03 $\pm$ 0.54 & 0.12 $\pm$ 0.01 \\ 
1 $\times$ [Bi]LSTM & 3 $\times$ [Bi]LSTM & 2 $\times$ [Bi]LSTM & 97.23 $\pm$ 0.41 & 97.23 $\pm$ 0.41 & 97.25 $\pm$ 0.38 & 97.23 $\pm$ 0.41 & 97.22 $\pm$ 0.43 & 0.08 $\pm$ 0.00 & 97.15 $\pm$ 0.42 & 97.15 $\pm$ 0.42 & 97.17 $\pm$ 0.40 & 97.15 $\pm$ 0.42 & 97.17 $\pm$ 0.42 & 0.14 $\pm$ 0.01 \\ 
1 $\times$ [Bi]LSTM & 3 $\times$ [Bi]LSTM & 3 $\times$ [Bi]LSTM & 97.43 $\pm$ 0.40 & 97.43 $\pm$ 0.40 & 97.45 $\pm$ 0.41 & 97.43 $\pm$ 0.40 & 97.42 $\pm$ 0.38 & 0.09 $\pm$ 0.00 & 97.24 $\pm$ 0.28 & 97.24 $\pm$ 0.28 & 97.24 $\pm$ 0.28 & 97.24 $\pm$ 0.28 & 97.28 $\pm$ 0.28 & 0.16 $\pm$ 0.01 \\ 
2 $\times$ [Bi]LSTM & 1 $\times$ [Bi]LSTM & 1 $\times$ [Bi]LSTM & \textbf{97.55 $\pm$ 0.38} & 97.55 $\pm$ 0.38 & 97.63 $\pm$ 0.37 & 97.55 $\pm$ 0.38 & 97.51 $\pm$ 0.38 & 0.10 $\pm$ 0.00 & \textbf{97.58 $\pm$ 0.73} & 97.58 $\pm$ 0.73 & 97.55 $\pm$ 0.65 & 97.58 $\pm$ 0.73 & 97.58 $\pm$ 0.70 & 0.18 $\pm$ 0.01 \\ 
2 $\times$ [Bi]LSTM & 1 $\times$ [Bi]LSTM & 2 $\times$ [Bi]LSTM & 97.43 $\pm$ 0.35 & 97.53 $\pm$ 0.35 & 97.48 $\pm$ 0.36 & 97.43 $\pm$ 0.35 & 97.42 $\pm$ 0.34 & 0.11 $\pm$ 0.00 & 97.18 $\pm$ 0.46 & 97.18 $\pm$ 0.46 & 97.24 $\pm$ 0.40 & 97.18 $\pm$ 0.46 & 97.16 $\pm$ 0.48 & 0.19 $\pm$ 0.01 \\ 
2 $\times$ [Bi]LSTM & 1 $\times$ [Bi]LSTM & 3 $\times$ [Bi]LSTM & 97.35 $\pm$ 0.47 & 97.35 $\pm$ 0.47 & 97.40 $\pm$ 0.46 & 97.35 $\pm$ 0.47 & 97.34 $\pm$ 0.45 & 0.13 $\pm$ 0.00 & 97.36 $\pm$ 0.27 & 97.36 $\pm$ 0.27 & 97.37 $\pm$ 0.28 & 97.36 $\pm$ 0.27 & 97.35 $\pm$ 0.26 & 0.22 $\pm$ 0.01 \\ 
2 $\times$ [Bi]LSTM & 2 $\times$ [Bi]LSTM & 1 $\times$ [Bi]LSTM & 97.00 $\pm$ 0.97 & 97.00 $\pm$ 0.97 & 97.16 $\pm$ 0.79 & 97.00 $\pm$ 0.97 & 93.97 $\pm$ 0.97 & 0.14 $\pm$ 0.00 & 97.04 $\pm$ 0.54 & 97.04 $\pm$ 0.54 & 97.13 $\pm$ 0.44 & 97.04 $\pm$ 0.54 & 97.05 $\pm$ 0.51 & 0.23 $\pm$ 0.01 \\ 
2 $\times$ [Bi]LSTM & 2 $\times$ [Bi]LSTM & 2 $\times$ [Bi]LSTM & 97.36 $\pm$ 0.57 & 97.36 $\pm$ 0.57 & 97.39 $\pm$ 0.54 & 97.36 $\pm$ 0.57 & 97.36 $\pm$ 0.57 & 0.15 $\pm$ 0.00 & 97.25 $\pm$ 0.39 & 97.25 $\pm$ 0.39 & 97.29 $\pm$ 0.35 & 97.25 $\pm$ 0.39 & 97.26 $\pm$ 0.39 & 0.25 $\pm$ 0.01 \\ 
2 $\times$ [Bi]LSTM & 2 $\times$ [Bi]LSTM & 3 $\times$ [Bi]LSTM & 97.23 $\pm$ 0.47 & 97.23 $\pm$ 0.47 & 97.29 $\pm$ 0.43 & 97.23 $\pm$ 0.47 & 97.23 $\pm$ 0.50 & 0.16 $\pm$ 0.00 & 97.23 $\pm$ 0.21 & 97.23 $\pm$ 0.21 & 97.26 $\pm$ 0.21 & 97.23 $\pm$ 0.21 & 97.24 $\pm$ 0.18 & 0.28 $\pm$ 0.01 \\ 
2 $\times$ [Bi]LSTM & 3 $\times$ [Bi]LSTM & 1 $\times$ [Bi]LSTM & 97.31 $\pm$ 0.60 & 97.31 $\pm$ 0.60 & 97.37 $\pm$ 0.58 & 97.31 $\pm$ 0.60 & 97.30 $\pm$ 0.59 & 0.18 $\pm$ 0.01 & 97.24 $\pm$ 0.44 & 97.24 $\pm$ 0.44 & 97.32 $\pm$ 0.37 & 97.24 $\pm$ 0.44 & 97.21 $\pm$ 0.47 & 0.30 $\pm$ 0.01 \\ 
2 $\times$ [Bi]LSTM & 3 $\times$ [Bi]LSTM & 2 $\times$ [Bi]LSTM & 97.47 $\pm$ 0.51 & 97.47 $\pm$ 0.51 & 97.53 $\pm$ 0.49 & 97.47 $\pm$ 0.51 & 97.45 $\pm$ 0.50 & 0.19 $\pm$ 0.01 & 97.26 $\pm$ 0.17 & 97.26 $\pm$ 0.17 & 97.31 $\pm$ 0.17 & 97.26 $\pm$ 0.17 & 97.24 $\pm$ 0.16 & 0.32 $\pm$ 0.01 \\ 
2 $\times$ [Bi]LSTM & 3 $\times$ [Bi]LSTM & 3 $\times$ [Bi]LSTM & 97.26 $\pm$ 0.43 & 97.26 $\pm$ 0.43 & 97.28 $\pm$ 0.41 & 97.26 $\pm$ 0.43 & 97.29 $\pm$ 0.43 & 0.21 $\pm$ 0.01 & 97.23 $\pm$ 0.31 & 97.23 $\pm$ 0.31 & 97.28 $\pm$ 0.25 & 97.23 $\pm$ 0.31 & 97.23 $\pm$ 0.35 & 0.35 $\pm$ 0.01 \\ 
3 $\times$ [Bi]LSTM & 1 $\times$ [Bi]LSTM & 1 $\times$ [Bi]LSTM & 97.40 $\pm$ 0.58 & 97.40 $\pm$ 0.58 & 97.45 $\pm$ 0.55 & 97.40 $\pm$ 0.58 & 97.39 $\pm$ 0.57 & 0.22 $\pm$ 0.01 & 97.40 $\pm$ 0.35 & 97.40 $\pm$ 0.35 & 97.40 $\pm$ 0.34 & 97.40 $\pm$ 0.35 & 97.43 $\pm$ 0.34 & 0.37 $\pm$ 0.01 \\ 
3 $\times$ [Bi]LSTM & 1 $\times$ [Bi]LSTM & 2 $\times$ [Bi]LSTM & 97.37 $\pm$ 0.47 & 97.37 $\pm$ 0.47 & 97.42 $\pm$ 0.48 & 97.37 $\pm$ 0.47 & 97.38 $\pm$ 0.45 & 0.23 $\pm$ 0.01 & 97.33 $\pm$ 0.34 & 97.33 $\pm$ 0.34 & 97.38 $\pm$ 0.32 & 97.33 $\pm$ 0.34 & 97.32 $\pm$ 0.32 & 0.39 $\pm$ 0.01 \\

3 $\times$ [Bi]LSTM & 1 $\times$ [Bi]LSTM & 3 $\times$ [Bi]LSTM & 97.41 $\pm$ 0.61 & 97.41 $\pm$ 0.61 & 97.44 $\pm$ 0.61 & 97.41 $\pm$ 0.61 & 97.41 $\pm$ 0.60 & 0.25 $\pm$ 0.01 & 97.35 $\pm$ 0.30 & 97.35 $\pm$ 0.30 & 97.37 $\pm$ 0.29 & 97.35 $\pm$ 0.30 & 97.35 $\pm$ 0.31 & 0.41 $\pm$ 0.01 \\ 
3 $\times$ [Bi]LSTM & 2 $\times$ [Bi]LSTM & 1 $\times$ [Bi]LSTM & 97.35 $\pm$ 0.41 & 97.35 $\pm$ 0.41 & 97.40 $\pm$ 0.40 & 97.35 $\pm$ 0.41 & 97.33 $\pm$ 0.40 & 0.26 $\pm$ 0.01 & 97.06 $\pm$ 0.79 & 97.06 $\pm$ 0.79 & 97.19 $\pm$ 0.57 & 97.06 $\pm$ 0.79 & 97.03 $\pm$ 0.85 & 0.43 $\pm$ 0.01 \\ 
3 $\times$ [Bi]LSTM & 2 $\times$ [Bi]LSTM & 2 $\times$ [Bi]LSTM & 97.25 $\pm$ 0.58 & 97.25 $\pm$ 0.58 & 97.29 $\pm$ 0.57 & 97.25 $\pm$ 0.58 & 97.27 $\pm$ 0.58 & 0.27 $\pm$ 0.01 & 97.35 $\pm$ 0.41 & 97.35 $\pm$ 0.41 & 97.37 $\pm$ 0.38 & 97.35 $\pm$ 0.41 & 97.33 $\pm$ 0.43 & 0.46 $\pm$ 0.01 \\ 
3 $\times$ [Bi]LSTM & 2 $\times$ [Bi]LSTM & 3 $\times$ [Bi]LSTM & 97.24 $\pm$ 0.60 & 97.24 $\pm$ 0.60 & 97.29 $\pm$ 0.59 & 97.24 $\pm$ 0.60 & 97.26 $\pm$ 0.56 & 0.29 $\pm$ 0.01 & 97.13 $\pm$ 0.32 & 97.13 $\pm$ 0.32 & 97.15 $\pm$ 0.30 & 97.13 $\pm$ 0.32 & 97.15 $\pm$ 0.31 & 0.48 $\pm$ 0.01 \\ 
3 $\times$ [Bi]LSTM & 3 $\times$ [Bi]LSTM & 1 $\times$ [Bi]LSTM & 97.47 $\pm$ 0.59 & 97.47 $\pm$ 0.59 & 97.52 $\pm$ 0.57 & 97.47 $\pm$ 0.59 & 97.47 $\pm$ 0.58 & 0.30 $\pm$ 0.01 & 97.34 $\pm$ 0.29 & 97.34 $\pm$ 0.29 & 97.35 $\pm$ 0.29 & 97.34 $\pm$ 0.29 & 97.36 $\pm$ 0.28 & 0.51 $\pm$ 0.02 \\ 
3 $\times$ [Bi]LSTM & 3 $\times$ [Bi]LSTM & 2 $\times$ [Bi]LSTM & 97.45 $\pm$ 0.53 & 97.45 $\pm$ 0.53 & 97.49 $\pm$ 0.52 & 97.45 $\pm$ 0.53 & 97.46 $\pm$ 0.51 & 0.32 $\pm$ 0.01 & 97.23 $\pm$ 0.44 & 97.23 $\pm$ 0.44 & 97.24 $\pm$ 0.43 & 97.23 $\pm$ 0.44 & 97.24 $\pm$ 0.41 & 0.53 $\pm$ 0.02 \\ 
3 $\times$ [Bi]LSTM & 3 $\times$ [Bi]LSTM & 3 $\times$ [Bi]LSTM & 97.48 $\pm$ 0.47 & 97.48 $\pm$ 0.47 & 97.49 $\pm$ 0.47 & 97.48 $\pm$ 0.47 & 97.48 $\pm$ 0.48 & 0.34 $\pm$ 0.01 & 97.23 $\pm$ 0.50 & 97.23 $\pm$ 0.50 & 97.27 $\pm$ 0.45 & 97.23 $\pm$ 0.50 & 97.23 $\pm$ 0.48 & 0.57 $\pm$ 0.02 \\ 
\hline
		\end{tabular}}
\end{table}
\vspace{-9pt}

\begin{table}[!htbp]
\caption{MisRoBÆRTa ablation and sensitivity testing using the \textsc{FakeNewsCorpus} dataset presented in Section~\ref{sec:results}. (Note: bold text marks the overall best result).} 
\label{tab:ablation}
\resizebox{\textwidth}{!}{%
\begin{tabular}{ccccccccccccccc}
\hline
\multicolumn{3}{c}{\textbf{32 Units/Layer}} & \multicolumn{6}{c}{\textbf{MisRoBÆRTa 
with LSTM Cells}} & \multicolumn{6}{c}{\textbf{MisRoBÆRTa with BiLSTM Cells}} \\ 
\hline
\textbf{\begin{tabular}[c]{@{}c@{}}BART\\ Branch\end{tabular}} & \textbf{\begin{tabular}[c]{@{}c@{}}RoBERTa\\ Branch\end{tabular}} & \textbf{\begin{tabular}[c]{@{}c@{}}Ensemble\\ Branch\end{tabular}} & \textbf{Accuracy} & \textbf{\begin{tabular}[c]{@{}c@{}}Precision\\ Micro\end{tabular}} & \textbf{\begin{tabular}[c]{@{}c@{}}Precision\\ Macro\end{tabular}} & \textbf{\begin{tabular}[c]{@{}c@{}}Recall\\ Micro\end{tabular}} & \textbf{\begin{tabular}[c]{@{}c@{}}Recall\\ Macro\end{tabular}} & \textbf{\begin{tabular}[c]{@{}c@{}}Execution\\ Time (Hours)\end{tabular}} & \textbf{Accuracy} & \textbf{\begin{tabular}[c]{@{}c@{}}Precision\\ Micro\end{tabular}} & \textbf{\begin{tabular}[c]{@{}c@{}}Precision\\ Macro\end{tabular}} & \textbf{\begin{tabular}[c]{@{}c@{}}Recall\\ Micro\end{tabular}} & \textbf{\begin{tabular}[c]{@{}c@{}}Recall\\ Macro\end{tabular}} & \textbf{\begin{tabular}[c]{@{}c@{}}Execution\\ Time (Hours)\end{tabular}} \\
\hline 
1 $\times$ [Bi]LSTM & 1 $\times$ [Bi]LSTM & 1 $\times$ [Bi]LSTM & 91.38 $\pm$ 0.24 & 91.38 $\pm$ 0.24 & 91.55 $\pm$ 0.26 & 91.38 $\pm$ 0.24 & 91.38 $\pm$ 0.24 & 0.03 $\pm$ 0.00 & 91.91 $\pm$ 0.30 & 91.91 $\pm$ 0.30 & 92.16 $\pm$ 0.20 & 91.91 $\pm$ 0.30 & 91.91 $\pm$ 0.30 & 0.03 $\pm$ 0.00 \\ 
1 $\times$ [Bi]LSTM & 1 $\times$ [Bi]LSTM & 2 $\times$ [Bi]LSTM & 91.35 $\pm$ 0.36 & 91.35 $\pm$ 0.36 & 91.57 $\pm$ 0.32 & 91.35 $\pm$ 0.36 & 91.35 $\pm$ 0.36 & 0.06 $\pm$ 0.00 & 91.90 $\pm$ 0.21 & 91.90 $\pm$ 0.21 & 92.08 $\pm$ 0.20 & 91.90 $\pm$ 0.21 & 91.90 $\pm$ 0.21 & 0.06 $\pm$ 0.00 \\ 
1 $\times$ [Bi]LSTM & 1 $\times$ [Bi]LSTM & 3 $\times$ [Bi]LSTM & 90.87 $\pm$ 0.34 & 90.87 $\pm$ 0.34 & 91.16 $\pm$ 0.35 & 90.87 $\pm$ 0.34 & 90.87 $\pm$ 0.34 & 0.11 $\pm$ 0.00 & 91.61 $\pm$ 0.21 & 91.61 $\pm$ 0.21 & 91.79 $\pm$ 0.17 & 91.61 $\pm$ 0.21 & 91.61 $\pm$ 0.21 & 0.10 $\pm$ 0.01 \\ 
1 $\times$ [Bi]LSTM & 2 $\times$ [Bi]LSTM & 1 $\times$ [Bi]LSTM & 90.90 $\pm$ 0.35 & 90.90 $\pm$ 0.35 & 91.13 $\pm$ 0.30 & 90.90 $\pm$ 0.35 & 90.90 $\pm$ 0.35 & 0.14 $\pm$ 0.01 & 91.62 $\pm$ 0.29 & 91.62 $\pm$ 0.29 & 91.84 $\pm$ 0.28 & 91.62 $\pm$ 0.29 & 91.62 $\pm$ 0.29 & 0.13 $\pm$ 0.01 \\ 
1 $\times$ [Bi]LSTM & 2 $\times$ [Bi]LSTM & 2 $\times$ [Bi]LSTM & 90.87 $\pm$ 0.25 & 90.87 $\pm$ 0.25 & 91.13 $\pm$ 0.25 & 90.87 $\pm$ 0.25 & 90.87 $\pm$ 0.25 & 0.18 $\pm$ 0.01 & 91.60 $\pm$ 0.16 & 91.60 $\pm$ 0.16 & 91.75 $\pm$ 0.17 & 91.60 $\pm$ 0.16 & 91.60 $\pm$ 0.16 & 0.17 $\pm$ 0.01 \\ 
1 $\times$ [Bi]LSTM & 2 $\times$ [Bi]LSTM & 3 $\times$ [Bi]LSTM & 90.60 $\pm$ 0.35 & 90.60 $\pm$ 0.35 & 90.79 $\pm$ 0.30 & 90.60 $\pm$ 0.35 & 90.60 $\pm$ 0.35 & 0.23 $\pm$ 0.01 & 91.45 $\pm$ 0.21 & 91.45 $\pm$ 0.21 & 91.60 $\pm$ 0.24 & 91.45 $\pm$ 0.21 & 91.45 $\pm$ 0.21 & 0.21 $\pm$ 0.01 \\ 
1 $\times$ [Bi]LSTM & 3 $\times$ [Bi]LSTM & 1 $\times$ [Bi]LSTM & 90.64 $\pm$ 0.43 & 90.64 $\pm$ 0.43 & 90.77 $\pm$ 0.43 & 90.64 $\pm$ 0.43 & 90.64 $\pm$ 0.43 & 0.26 $\pm$ 0.01 & 91.51 $\pm$ 0.22 & 91.51 $\pm$ 0.22 & 91.70 $\pm$ 0.24 & 91.51 $\pm$ 0.22 & 91.51 $\pm$ 0.22 & 0.25 $\pm$ 0.01 \\ 
1 $\times$ [Bi]LSTM & 3 $\times$ [Bi]LSTM & 2 $\times$ [Bi]LSTM & 90.83 $\pm$ 0.21 & 90.83 $\pm$ 0.21 & 91.09 $\pm$ 0.18 & 90.83 $\pm$ 0.21 & 90.83 $\pm$ 0.21 & 0.31 $\pm$ 0.01 & 91.60 $\pm$ 0.21 & 91.60 $\pm$ 0.21 & 91.78 $\pm$ 0.24 & 91.60 $\pm$ 0.21 & 91.59 $\pm$ 0.21 & 0.29 $\pm$ 0.01 \\ 
1 $\times$ [Bi]LSTM & 3 $\times$ [Bi]LSTM & 3 $\times$ [Bi]LSTM & 90.56 $\pm$ 0.34 & 90.56 $\pm$ 0.34 & 90.79 $\pm$ 0.32 & 90.56 $\pm$ 0.34 & 90.56 $\pm$ 0.34 & 0.36 $\pm$ 0.02 & 91.32 $\pm$ 0.18 & 91.32 $\pm$ 0.18 & 91.53 $\pm$ 0.16 & 91.32 $\pm$ 0.18 & 91.32 $\pm$ 0.18 & 0.35 $\pm$ 0.02 \\ 
2 $\times$ [Bi]LSTM & 1 $\times$ [Bi]LSTM & 1 $\times$ [Bi]LSTM & \textbf{91.98} $\pm$ 0.19 & 91.98 $\pm$ 0.19 & 92.17$\pm$ 0.15 & 91.98 $\pm$ 0.19 & 91.98 $\pm$ 0.19 & 0.39 $\pm$ 0.02 & \textbf{92.08} $\pm$ 0.18 & 92.08 $\pm$ 0.18 & 92.25 $\pm$ 0.19 & 92.08 $\pm$ 0.18 & 92.08 $\pm$ 0.18 & 0.38 $\pm$ 0.02 \\ 
2 $\times$ [Bi]LSTM & 1 $\times$ [Bi]LSTM & 2 $\times$ [Bi]LSTM & 91.60 $\pm$ 0.19 & 91.60 $\pm$ 0.19 & 91.78 $\pm$ 0.25 & 91.60 $\pm$ 0.19 & 91.60 $\pm$ 0.19 & 0.43 $\pm$ 0.02 & 91.80 $\pm$ 0.17 & 91.80 $\pm$ 0.17 & 91.94 $\pm$ 0.18 & 91.80 $\pm$ 0.17 & 91.80 $\pm$ 0.17 & 0.42 $\pm$ 0.02 \\ 
2 $\times$ [Bi]LSTM & 1 $\times$ [Bi]LSTM & 3 $\times$ [Bi]LSTM & 91.19 $\pm$ 0.28 & 91.19 $\pm$ 0.28 & 91.37 $\pm$ 0.27 & 91.19 $\pm$ 0.28 & 91.19 $\pm$ 0.28 & 0.48 $\pm$ 0.02 & 91.53 $\pm$ 0.30 & 91.53 $\pm$ 0.30 & 91.67 $\pm$ 0.30 & 91.53 $\pm$ 0.30 & 91.53 $\pm$ 0.30 & 0.47 $\pm$ 0.02 \\ 
2 $\times$ [Bi]LSTM & 2 $\times$ [Bi]LSTM & 1 $\times$ [Bi]LSTM & 91.73 $\pm$ 0.12 & 91.73 $\pm$ 0.12 & 91.90 $\pm$ 0.09 & 91.73 $\pm$ 0.12 & 91.73 $\pm$ 0.12 & 0.51 $\pm$ 0.02 & 91.88 $\pm$ 0.17 & 91.88 $\pm$ 0.17 & 92.07 $\pm$ 0.20 & 91.88 $\pm$ 0.17 & 91.88 $\pm$ 0.17 & 0.51 $\pm$ 0.02 \\ 
2 $\times$ [Bi]LSTM & 2 $\times$ [Bi]LSTM & 2 $\times$ [Bi]LSTM & 91.35 $\pm$ 0.24 & 91.35 $\pm$ 0.24 & 91.51 $\pm$ 0.22 & 91.35 $\pm$ 0.24 & 91.35 $\pm$ 0.24 & 0.56 $\pm$ 0.02 & 91.71 $\pm$ 0.20 & 91.71 $\pm$ 0.20 & 91.86 $\pm$ 0.19 & 91.71 $\pm$ 0.20 & 91.71 $\pm$ 0.20 & 0.56 $\pm$ 0.02 \\ 
2 $\times$ [Bi]LSTM & 2 $\times$ [Bi]LSTM & 3 $\times$ [Bi]LSTM & 90.92 $\pm$ 0.37 & 90.92 $\pm$ 0.37 & 91.15 $\pm$ 0.34 & 90.92 $\pm$ 0.37 & 90.92 $\pm$ 0.37 & 0.61 $\pm$ 0.02 & 91.20 $\pm$ 0.25 & 91.20 $\pm$ 0.25 & 91.32 $\pm$ 0.24 & 91.20 $\pm$ 0.25 & 91.20 $\pm$ 0.25 & 0.62 $\pm$ 0.03 \\ 
2 $\times$ [Bi]LSTM & 3 $\times$ [Bi]LSTM & 1 $\times$ [Bi]LSTM & 91.55 $\pm$ 0.18 & 91.55 $\pm$ 0.18 & 91.68 $\pm$ 0.21 & 91.55 $\pm$ 0.18 & 91.55 $\pm$ 0.18 & 0.64 $\pm$ 0.02 & 91.56 $\pm$ 0.21 & 91.56 $\pm$ 0.21 & 91.75 $\pm$ 0.19 & 91.56 $\pm$ 0.21 & 91.56 $\pm$ 0.21 & 0.67 $\pm$ 0.03 \\ 
2 $\times$ [Bi]LSTM & 3 $\times$ [Bi]LSTM & 2 $\times$ [Bi]LSTM & 91.10 $\pm$ 0.46 & 91.10 $\pm$ 0.46 & 91.23 $\pm$ 0.44 & 91.10 $\pm$ 0.46 & 91.10 $\pm$ 0.46 & 0.69 $\pm$ 0.03 & 91.49 $\pm$ 0.21 & 91.49 $\pm$ 0.21 & 91.58 $\pm$ 0.24 & 91.49 $\pm$ 0.21 & 91.49 $\pm$ 0.21 & 0.72 $\pm$ 0.03 \\ 
2 $\times$ [Bi]LSTM & 3 $\times$ [Bi]LSTM & 3 $\times$ [Bi]LSTM & 90.69 $\pm$ 0.37 & 90.69 $\pm$ 0.37 & 90.92 $\pm$ 0.34 & 90.69 $\pm$ 0.37 & 90.69 $\pm$ 0.37 & 0.75 $\pm$ 0.03 & 91.19 $\pm$ 0.25 & 91.19 $\pm$ 0.25 & 91.35 $\pm$ 0.32 & 91.19 $\pm$ 0.25 & 91.19 $\pm$ 0.25 & 0.79 $\pm$ 0.03 \\ 
3 $\times$ [Bi]LSTM & 1 $\times$ [Bi]LSTM & 1 $\times$ [Bi]LSTM & 91.74 $\pm$ 0.23 & 91.74 $\pm$ 0.23 & 91.92 $\pm$ 0.22 & 91.74 $\pm$ 0.23 & 91.74 $\pm$ 0.23 & 0.80 $\pm$ 0.03 & 91.92 $\pm$ 0.22 & 91.92 $\pm$ 0.22 & 92.06 $\pm$ 0.23 & 91.92 $\pm$ 0.22 & 91.92 $\pm$ 0.22 & 0.84 $\pm$ 0.04 \\ 
3 $\times$ [Bi]LSTM & 1 $\times$ [Bi]LSTM & 2 $\times$ [Bi]LSTM & 91.39 $\pm$ 0.20 & 91.39 $\pm$ 0.20 & 91.58 $\pm$ 0.17 & 91.39 $\pm$ 0.20 & 91.39 $\pm$ 0.20 & 0.85 $\pm$ 0.03 & 91.72 $\pm$ 0.32 & 91.72 $\pm$ 0.32 & 91.91 $\pm$ 0.31 & 91.72 $\pm$ 0.32 & 91.72 $\pm$ 0.32 & 0.92 $\pm$ 0.04 \\ 
3 $\times$ [Bi]LSTM & 1 $\times$ [Bi]LSTM & 3 $\times$ [Bi]LSTM & 90.86 $\pm$ 0.20 & 90.86 $\pm$ 0.20 & 91.06 $\pm$ 0.16 & 90.86 $\pm$ 0.20 & 90.86 $\pm$ 0.20 & 0.90 $\pm$ 0.04 & 91.48 $\pm$ 0.26 & 91.48 $\pm$ 0.26 & 91.64 $\pm$ 0.23 & 91.48 $\pm$ 0.26 & 91.48 $\pm$ 0.26 & 0.97 $\pm$ 0.04 \\ 
3 $\times$ [Bi]LSTM & 2 $\times$ [Bi]LSTM & 1 $\times$ [Bi]LSTM & 91.62 $\pm$ 0.19 & 91.62 $\pm$ 0.19 & 91.82 $\pm$ 0.17 & 91.62 $\pm$ 0.19 & 91.62 $\pm$ 0.19 & 0.95 $\pm$ 0.04 & 91.84 $\pm$ 0.25 & 91.84 $\pm$ 0.25 & 92.00 $\pm$ 0.28 & 91.84 $\pm$ 0.25 & 91.84 $\pm$ 0.25 & 1.02 $\pm$ 0.04 \\ 
3 $\times$ [Bi]LSTM & 2 $\times$ [Bi]LSTM & 2 $\times$ [Bi]LSTM & 91.16 $\pm$ 0.26 & 91.16 $\pm$ 0.26 & 91.31 $\pm$ 0.26 & 91.16 $\pm$ 0.26 & 91.16 $\pm$ 0.26 & 1.01 $\pm$ 0.04 & 91.51 $\pm$ 0.19 & 91.51 $\pm$ 0.19 & 91.66 $\pm$ 0.17 & 91.51 $\pm$ 0.19 & 91.51 $\pm$ 0.19 & 1.09 $\pm$ 0.05 \\ 
3 $\times$ [Bi]LSTM & 2 $\times$ [Bi]LSTM & 3 $\times$ [Bi]LSTM & 90.36 $\pm$ 0.00 & 90.36 $\pm$ 0.00 & 90.79 $\pm$ 0.00 & 90.36 $\pm$ 0.00 & 90.36 $\pm$ 0.00 & 1.03 $\pm$ 0.04 & 91.25 $\pm$ 0.27 & 91.25 $\pm$ 0.27 & 91.49 $\pm$ 0.27 & 91.25 $\pm$ 0.27 & 91.25 $\pm$ 0.27 & 1.17 $\pm$ 0.06 \\ 
3 $\times$ [Bi]LSTM & 3 $\times$ [Bi]LSTM & 1 $\times$ [Bi]LSTM & 91.23 $\pm$ 0.28 & 91.23 $\pm$ 0.28 & 91.37 $\pm$ 0.26 & 91.23 $\pm$ 0.28 & 91.23 $\pm$ 0.28 & 1.09 $\pm$ 0.04 & 91.50 $\pm$ 0.23 & 91.50 $\pm$ 0.23 & 91.63 $\pm$ 0.21 & 91.50 $\pm$ 0.23 & 91.50 $\pm$ 0.23 & 1.23 $\pm$ 0.06 \\ 
3 $\times$ [Bi]LSTM & 3 $\times$ [Bi]LSTM & 2 $\times$ [Bi]LSTM & 90.77 $\pm$ 0.28 & 90.77 $\pm$ 0.28 & 90.92 $\pm$ 0.30 & 90.77 $\pm$ 0.28 & 90.77 $\pm$ 0.28 & 1.15 $\pm$ 0.04 & 91.32 $\pm$ 0.19 & 91.32 $\pm$ 0.19 & 91.47 $\pm$ 0.18 & 91.32 $\pm$ 0.19 & 91.32 $\pm$ 0.19 & 1.30 $\pm$ 0.06 \\ 
3 $\times$ [Bi]LSTM & 3 $\times$ [Bi]LSTM & 3 $\times$ [Bi]LSTM & 88.50 $\pm$ 0.36 & 88.50 $\pm$ 0.36 & 88.87 $\pm$ 0.24 & 88.50 $\pm$ 0.36 & 88.50 $\pm$ 0.36 & 1.18 $\pm$ 0.05 & 91.22 $\pm$ 0.25 & 91.22 $\pm$ 0.25 & 91.41 $\pm$ 0.20 & 91.22 $\pm$ 0.25 & 91.22 $\pm$ 0.25 & 1.36 $\pm$ 0.08 \\  

\multicolumn{3}{c}{\textbf{64 Units/Layer}} & \multicolumn{6}{c}{\textbf{MisRoBÆRTa with LSTM Cells}} & \multicolumn{6}{c}{\textbf{MisRoBÆRTa with BiLSTM Cells}} \\ 
\hline
\textbf{\begin{tabular}[c]{@{}c@{}}BART\\ Branch\end{tabular}} & \textbf{\begin{tabular}[c]{@{}c@{}}RoBERTa\\ Branch\end{tabular}} & \textbf{\begin{tabular}[c]{@{}c@{}}Ensemble\\ Branch\end{tabular}} & \textbf{Accuracy} & \textbf{\begin{tabular}[c]{@{}c@{}}Precision\\ Micro\end{tabular}} & \textbf{\begin{tabular}[c]{@{}c@{}}Precision\\ Macro\end{tabular}} & \textbf{\begin{tabular}[c]{@{}c@{}}Recall\\ Micro\end{tabular}} & \textbf{\begin{tabular}[c]{@{}c@{}}Recall\\ Macro\end{tabular}} & \textbf{\begin{tabular}[c]{@{}c@{}}Execution\\ Time (Hours)\end{tabular}} & \textbf{Accuracy} & \textbf{\begin{tabular}[c]{@{}c@{}}Precision\\ Micro\end{tabular}} & \textbf{\begin{tabular}[c]{@{}c@{}}Precision\\ Macro\end{tabular}} & \textbf{\begin{tabular}[c]{@{}c@{}}Recall\\ Micro\end{tabular}} & \textbf{\begin{tabular}[c]{@{}c@{}}Recall\\ Macro\end{tabular}} & \textbf{\begin{tabular}[c]{@{}c@{}}Execution\\ Time (Hours)\end{tabular}} \\ 
\hline
1 $\times$ [Bi]LSTM & 1 $\times$ [Bi]LSTM & 1 $\times$ [Bi]LSTM & 92.05 $\pm$ 0.18 & 92.05 $\pm$ 0.18 & 92.23 $\pm$ 0.19 & 92.05 $\pm$ 0.18 & 92.05 $\pm$ 0.18 & 0.02 $\pm$ 0.00 & 92.26 $\pm$ 0.23 & 92.26 $\pm$ 0.23 & 92.40 $\pm$ 0.21 & 92.26 $\pm$ 0.23 & 92.26 $\pm$ 0.23 & 0.02 $\pm$ 0.00 \\ 
1 $\times$ [Bi]LSTM & 1 $\times$ [Bi]LSTM & 2 $\times$ [Bi]LSTM & 91.77 $\pm$ 0.29 & 91.77 $\pm$ 0.29 & 91.96 $\pm$ 0.25 & 91.77 $\pm$ 0.29 & 91.77 $\pm$ 0.29 & 0.04 $\pm$ 0.00 & 92.12 $\pm$ 0.19 & 92.12 $\pm$ 0.19 & 92.22 $\pm$ 0.19 & 92.12 $\pm$ 0.19 & 92.12 $\pm$ 0.19 & 0.05 $\pm$ 0.00 \\ 
1 $\times$ [Bi]LSTM & 1 $\times$ [Bi]LSTM & 3 $\times$ [Bi]LSTM & 91.72 $\pm$ 0.21 & 91.72 $\pm$ 0.21 & 91.90 $\pm$ 0.21 & 91.72 $\pm$ 0.21 & 91.72 $\pm$ 0.21 & 0.07 $\pm$ 0.00 & 91.90 $\pm$ 0.17 & 91.90 $\pm$ 0.17 & 92.07 $\pm$ 0.18 & 91.90 $\pm$ 0.17 & 91.90 $\pm$ 0.17 & {0.08 $\pm$ 0.00} \\ 
1 $\times$ [Bi]LSTM & 2 $\times$ [Bi]LSTM & 1 $\times$ [Bi]LSTM & 91.72 $\pm$ 0.28 & 91.72 $\pm$ 0.28 & 91.92 $\pm$ 0.29 & 91.72 $\pm$ 0.28 & 91.72 $\pm$ 0.28 & 0.09 $\pm$ 0.00 & 92.02 $\pm$ 0.20 & 92.02 $\pm$ 0.20 & 92.16 $\pm$ 0.18 & 92.02 $\pm$ 0.20 & 92.02 $\pm$ 0.20 & {0.11 $\pm$ 0.01} \\ 
1 $\times$ [Bi]LSTM & 2 $\times$ [Bi]LSTM & 2 $\times$ [Bi]LSTM & 91.63 $\pm$ 0.13 & 91.63 $\pm$ 0.13 & 91.83 $\pm$ 0.13 & 91.63 $\pm$ 0.13 & 91.63 $\pm$ 0.13 & 0.11 $\pm$ 0.00 & 91.92 $\pm$ 0.28 & 91.92 $\pm$ 0.28 & 92.14 $\pm$ 0.21 & 91.92 $\pm$ 0.28 & 91.92 $\pm$ 0.28 & {0.14 $\pm$ 0.01} \\ 
1 $\times$ [Bi]LSTM & 2 $\times$ [Bi]LSTM & 3 $\times$ [Bi]LSTM & 91.48 $\pm$ 0.35 & 91.48 $\pm$ 0.35 & 91.73 $\pm$ 0.28 & 91.48 $\pm$ 0.35 & 91.48 $\pm$ 0.35 & 0.15 $\pm$ 0.01 & 91.97 $\pm$ 0.15 & 91.97 $\pm$ 0.15 & 92.06 $\pm$ 0.15 & 91.97 $\pm$ 0.15 & 91.97 $\pm$ 0.15 & {0.18 $\pm$ 0.01} \\ 
1 $\times$ [Bi]LSTM & 3 $\times$ [Bi]LSTM & 1 $\times$ [Bi]LSTM & 91.57 $\pm$ 0.23 & 91.57 $\pm$ 0.23 & 91.78 $\pm$ 0.20 & 91.57 $\pm$ 0.23 & 91.57 $\pm$ 0.23 & 0.17 $\pm$ 0.01 & 92.01 $\pm$ 0.16 & 92.01 $\pm$ 0.16 & 92.16 $\pm$ 0.19 & 92.01 $\pm$ 0.16 & 92.01 $\pm$ 0.16 & {0.21 $\pm$ 0.01} \\ 
1 $\times$ [Bi]LSTM & 3 $\times$ [Bi]LSTM & 2 $\times$ [Bi]LSTM & 91.66 $\pm$ 0.16 & 91.66 $\pm$ 0.16 & 91.79 $\pm$ 0.18 & 91.66 $\pm$ 0.16 & 91.66 $\pm$ 0.16 & 0.19 $\pm$ 0.01 & 91.90 $\pm$ 0.21 & 91.90 $\pm$ 0.21 & 92.08 $\pm$ 0.22 & 91.90 $\pm$ 0.21 & 91.90 $\pm$ 0.21 & {0.25 $\pm$ 0.01} \\ 
1 $\times$ [Bi]LSTM & 3 $\times$ [Bi]LSTM & 3 $\times$ [Bi]LSTM & 91.37 $\pm$ 0.26 & 91.37 $\pm$ 0.26 & 91.55 $\pm$ 0.27 & 91.37 $\pm$ 0.26 & 91.37 $\pm$ 0.26 & 0.23 $\pm$ 0.01 & 91.64 $\pm$ 0.22 & 91.64 $\pm$ 0.22 & 91.77 $\pm$ 0.24 & 91.64 $\pm$ 0.22 & 91.64 $\pm$ 0.22 & {0.29 $\pm$ 0.01} \\ 
2 $\times$ [Bi]LSTM & 1 $\times$ [Bi]LSTM & 1 $\times$ [Bi]LSTM & \textbf{92.30} $\pm$ 0.30 & 92.30 $\pm$ 0.30 & 92.47 $\pm$ 0.24 & 92.30 $\pm$ 0.30 & 92.30 $\pm$ 0.30 & 0.25 $\pm$ 0.01 & \textbf{92.41} $\pm$ 0.18 & 92.41 $\pm$ 0.18 & 92.57 $\pm$ 0.13 & 92.41 $\pm$ 0.18 & 92.41 $\pm$ 0.18 & {0.32 $\pm$ 0.01} \\ 
2 $\times$ [Bi]LSTM & 1 $\times$ [Bi]LSTM & 2 $\times$ [Bi]LSTM & 92.00 $\pm$ 0.35 & 92.00 $\pm$ 0.35 & 92.14 $\pm$ 0.32 & 92.00 $\pm$ 0.35 & 92.00 $\pm$ 0.35 & 0.28 $\pm$ 0.01 & 92.22 $\pm$ 0.17 & 92.22 $\pm$ 0.17 & 92.40 $\pm$ 0.16 & 92.22 $\pm$ 0.17 & 92.22 $\pm$ 0.17 & {0.35 $\pm$ 0.01} \\ 
2 $\times$ [Bi]LSTM & 1 $\times$ [Bi]LSTM & 3 $\times$ [Bi]LSTM & 91.94 $\pm$ 0.29 & 91.94 $\pm$ 0.29 & 92.06 $\pm$ 0.31 & 91.94 $\pm$ 0.29 & 91.94 $\pm$ 0.29 & 0.31 $\pm$ 0.01 & 92.12 $\pm$ 0.27 & 92.12 $\pm$ 0.27 & 92.24 $\pm$ 0.25 & 92.12 $\pm$ 0.27 & 92.12 $\pm$ 0.27 & {0.40 $\pm$ 0.01} \\ 
2 $\times$ [Bi]LSTM & 2 $\times$ [Bi]LSTM & 1 $\times$ [Bi]LSTM & 92.06 $\pm$ 0.21 & 92.06 $\pm$ 0.21 & 92.18 $\pm$ 0.23 & 92.06 $\pm$ 0.21 & 92.06 $\pm$ 0.21 & 0.33 $\pm$ 0.01 & 92.32 $\pm$ 0.23 & 92.32 $\pm$ 0.23 & 92.49 $\pm$ 0.24 & 92.32 $\pm$ 0.23 & 92.32 $\pm$ 0.23 & {0.43 $\pm$ 0.02} \\ 
2 $\times$ [Bi]LSTM & 2 $\times$ [Bi]LSTM & 2 $\times$ [Bi]LSTM & 91.83 $\pm$ 0.25 & 91.83 $\pm$ 0.25 & 91.99 $\pm$ 0.25 & 91.83 $\pm$ 0.25 & 91.83 $\pm$ 0.25 & 0.36 $\pm$ 0.01 & 92.16 $\pm$ 0.25 & 92.16 $\pm$ 0.25 & 92.31 $\pm$ 0.15 & 92.16 $\pm$ 0.25 & 92.16 $\pm$ 0.25 & {0.47 $\pm$ 0.02} \\ 
2 $\times$ [Bi]LSTM & 2 $\times$ [Bi]LSTM & 3 $\times$ [Bi]LSTM & 91.61 $\pm$ 0.19 & 91.61 $\pm$ 0.19 & 91.77 $\pm$ 0.21 & 91.61 $\pm$ 0.19 & 91.61 $\pm$ 0.19 & 0.40 $\pm$ 0.01 & 91.74 $\pm$ 0.25 & 91.74 $\pm$ 0.25 & 91.91 $\pm$ 0.26 & 91.74 $\pm$ 0.25 & 91.74 $\pm$ 0.25 & {0.52 $\pm$ 0.02} \\ 
2 $\times$ [Bi]LSTM & 3 $\times$ [Bi]LSTM & 1 $\times$ [Bi]LSTM & 91.97 $\pm$ 0.27 & 91.97 $\pm$ 0.27 & 92.10 $\pm$ 0.24 & 91.97 $\pm$ 0.27 & 91.97 $\pm$ 0.27 & 0.43 $\pm$ 0.01 & 92.01 $\pm$ 0.25 & 92.01 $\pm$ 0.25 & 92.19 $\pm$ 0.16 & 92.01 $\pm$ 0.25 & 92.01 $\pm$ 0.25 & {0.56 $\pm$ 0.02} \\ 
2 $\times$ [Bi]LSTM & 3 $\times$ [Bi]LSTM & 2 $\times$ [Bi]LSTM & 91.81 $\pm$ 0.31 & 91.81 $\pm$ 0.31 & 91.97 $\pm$ 0.31 & 91.81 $\pm$ 0.31 & 91.81 $\pm$ 0.31 & 0.46 $\pm$ 0.02 & 91.81 $\pm$ 0.17 & 91.81 $\pm$ 0.17 & 91.96 $\pm$ 0.23 & 91.81 $\pm$ 0.17 & 91.81 $\pm$ 0.17 & {0.60 $\pm$ 0.02} \\ 
2 $\times$ [Bi]LSTM & 3 $\times$ [Bi]LSTM & 3 $\times$ [Bi]LSTM & 91.47 $\pm$ 0.26 & 91.47 $\pm$ 0.26 & 91.64 $\pm$ 0.24 & 91.47 $\pm$ 0.26 & 91.47 $\pm$ 0.26 & 0.50 $\pm$ 0.02 & 91.51 $\pm$ 0.43 & 91.51 $\pm$ 0.43 & 91.70 $\pm$ 0.41 & 91.51 $\pm$ 0.43 & 91.51 $\pm$ 0.43 & {0.66 $\pm$ 0.02} \\ 
3 $\times$ [Bi]LSTM & 1 $\times$ [Bi]LSTM & 1 $\times$ [Bi]LSTM & 92.29 $\pm$ 0.26 & 92.29 $\pm$ 0.26 & 92.43 $\pm$ 0.21 & 92.29 $\pm$ 0.26 & 92.29 $\pm$ 0.26 & 0.52 $\pm$ 0.02 & 92.19 $\pm$ 0.31 & 92.19 $\pm$ 0.31 & 92.31 $\pm$ 0.34 & 92.19 $\pm$ 0.31 & 92.19 $\pm$ 0.31 & {0.70 $\pm$ 0.02} \\ 
3 $\times$ [Bi]LSTM & 1 $\times$ [Bi]LSTM & 2 $\times$ [Bi]LSTM & 91.90 $\pm$ 0.21 & 91.90 $\pm$ 0.21 & 92.09 $\pm$ 0.16 & 91.90 $\pm$ 0.21 & 91.90 $\pm$ 0.21 & 0.56 $\pm$ 0.02 & 91.97 $\pm$ 0.21 & 91.97 $\pm$ 0.21 & 92.09 $\pm$ 0.20 & 91.97 $\pm$ 0.21 & 91.97 $\pm$ 0.21 & {0.74 $\pm$ 0.03} \\ 
3 $\times$ [Bi]LSTM & 1 $\times$ [Bi]LSTM & 3 $\times$ [Bi]LSTM & 91.85 $\pm$ 0.28 & 91.85 $\pm$ 0.28 & 91.98 $\pm$ 0.26 & 91.85 $\pm$ 0.28 & 91.85 $\pm$ 0.28 & 0.60 $\pm$ 0.02 & 91.91 $\pm$ 0.15 & 91.91 $\pm$ 0.15 & 92.06 $\pm$ 0.17 & 91.91 $\pm$ 0.15 & 91.91 $\pm$ 0.15 & {0.80 $\pm$ 0.03} \\ 
3 $\times$ [Bi]LSTM & 2 $\times$ [Bi]LSTM & 1 $\times$ [Bi]LSTM & 92.08 $\pm$ 0.19 & 92.08 $\pm$ 0.19 & 92.20 $\pm$ 0.19 & 92.08 $\pm$ 0.19 & 92.08 $\pm$ 0.19 & 0.63 $\pm$ 0.02 & 92.16 $\pm$ 0.23 & 92.16 $\pm$ 0.23 & 92.31 $\pm$ 0.19 & 92.16 $\pm$ 0.23 & 92.16 $\pm$ 0.23 & {0.84 $\pm$ 0.03} \\ 
3 $\times$ [Bi]LSTM & 2 $\times$ [Bi]LSTM & 2 $\times$ [Bi]LSTM & 91.73 $\pm$ 0.27 & 91.73 $\pm$ 0.27 & 91.93 $\pm$ 0.27 & 91.73 $\pm$ 0.27 & 91.73 $\pm$ 0.27 & 0.67 $\pm$ 0.02 & 91.91 $\pm$ 0.13 & 91.91 $\pm$ 0.13 & 92.03 $\pm$ 0.16 & 91.90 $\pm$ 0.13 & 91.91 $\pm$ 0.13 & 0.92 $\pm$ 0.03 \\ 
3 $\times$ [Bi]LSTM & 2 $\times$ [Bi]LSTM & 3 $\times$ [Bi]LSTM & 91.43 $\pm$ 0.16 & 91.43 $\pm$ 0.16 & 91.63 $\pm$ 0.16 & 91.43 $\pm$ 0.16 & 91.43 $\pm$ 0.16 & 0.70 $\pm$ 0.02 & 91.59 $\pm$ 0.07 & 91.59 $\pm$ 0.07 & 91.71 $\pm$ 0.11 & 91.59 $\pm$ 0.07 & 91.59 $\pm$ 0.07 & 0.93 $\pm$ 0.04 \\ 
3 $\times$ [Bi]LSTM & 3 $\times$ [Bi]LSTM & 1 $\times$ [Bi]LSTM & 91.92 $\pm$ 0.22 & 91.92 $\pm$ 0.22 & 92.03 $\pm$ 0.22 & 91.92 $\pm$ 0.22 & 91.92 $\pm$ 0.22 & 0.73 $\pm$ 0.02 & 91.97 $\pm$ 0.23 & 91.97 $\pm$ 0.23 & 92.13 $\pm$ 0.24 & 91.97 $\pm$ 0.23 & 91.97 $\pm$ 0.23 & 0.98 $\pm$ 0.04 \\ 
3 $\times$ [Bi]LSTM & 3 $\times$ [Bi]LSTM & 2 $\times$ [Bi]LSTM & 91.61 $\pm$ 0.28 & 91.61 $\pm$ 0.28 & 91.77 $\pm$ 0.27 & 91.61 $\pm$ 0.28 & 91.61 $\pm$ 0.28 & 0.77 $\pm$ 0.03 & 91.80 $\pm$ 0.16 & 91.80 $\pm$ 0.16 & 91.93 $\pm$ 0.18 & 91.81 $\pm$ 0.16 & 91.80 $\pm$ 0.16 & 1.04 $\pm$ 0.04 \\ 
3 $\times$ [Bi]LSTM & 3 $\times$ [Bi]LSTM & 3 $\times$ [Bi]LSTM & 89.81 $\pm$ 0.21 & 89.81 $\pm$ 0.21 & 89.91 $\pm$ 0.27 & 89.81 $\pm$ 0.21 & 89.81 $\pm$ 0.21 & 0.79 $\pm$ 0.03 & 85.53 $\pm$ 0.11 & 85.53 $\pm$ 0.11 & 85.64 $\pm$ 0.17 & 85.53 $\pm$ 0.11 & 85.53 $\pm$ 0.11 & 1.08 $\pm$ 0.04 \\

\hline
\multicolumn{3}{c}{\textbf{128 Units/Layer}} & \multicolumn{6}{c}{\textbf{MisRoBÆRTa with LSTM Cells}} & \multicolumn{6}{c}{\textbf{MisRoBÆRTa with BiLSTM Cells}} \\ 
\hline
\textbf{\begin{tabular}[c]{@{}c@{}}BART\\ Branch\end{tabular}} & \textbf{\begin{tabular}[c]{@{}c@{}}RoBERTa\\ Branch\end{tabular}} & \textbf{\begin{tabular}[c]{@{}c@{}}Ensemble\\ Branch\end{tabular}} & \textbf{Accuracy} & \textbf{\begin{tabular}[c]{@{}c@{}}Precision\\ Micro\end{tabular}} & \textbf{\begin{tabular}[c]{@{}c@{}}Precision\\ Macro\end{tabular}} & \textbf{\begin{tabular}[c]{@{}c@{}}Recall\\ Micro\end{tabular}} & \textbf{\begin{tabular}[c]{@{}c@{}}Recall\\ Macro\end{tabular}} & \textbf{\begin{tabular}[c]{@{}c@{}}Execution\\ Time (Hours)\end{tabular}} & \textbf{Accuracy} & \textbf{\begin{tabular}[c]{@{}c@{}}Precision\\ Micro\end{tabular}} & \textbf{\begin{tabular}[c]{@{}c@{}}Precision\\ Macro\end{tabular}} & \textbf{\begin{tabular}[c]{@{}c@{}}Recall\\ Micro\end{tabular}} & \textbf{\begin{tabular}[c]{@{}c@{}}Recall\\ Macro\end{tabular}} & \textbf{\begin{tabular}[c]{@{}c@{}}Execution\\ Time (Hours)\end{tabular}} \\ 
\hline
1 $\times$ [Bi]LSTM & 1 $\times$ [Bi]LSTM & 1 $\times$ [Bi]LSTM & 92.16 $\pm$ 0.27 & 92.16 $\pm$ 0.27 & 92.35 $\pm$ 0.22 & 92.16 $\pm$ 0.27 & 92.16 $\pm$ 0.27 & 0.02 $\pm$ 0.00 & 92.20 $\pm$ 0.30 & 92.20 $\pm$ 0.30 & 92.35 $\pm$ 0.29 & 92.20 $\pm$ 0.30 & 92.20 $\pm$ 0.30 & 0.03 $\pm$ 0.00 \\ 
1 $\times$ [Bi]LSTM & 1 $\times$ [Bi]LSTM & 2 $\times$ [Bi]LSTM & 92.11 $\pm$ 0.37 & 92.11 $\pm$ 0.37 & 92.26 $\pm$ 0.35 & 92.11 $\pm$ 0.37 & 92.11 $\pm$ 0.37 & 0.03 $\pm$ 0.00 & 92.23 $\pm$ 0.14 & 92.23 $\pm$ 0.14 & 92.37 $\pm$ 0.13 & 92.23 $\pm$ 0.14 & 92.23 $\pm$ 0.14 & 0.05 $\pm$ 0.00 \\ 
1 $\times$ [Bi]LSTM & 1 $\times$ [Bi]LSTM & 3 $\times$ [Bi]LSTM & 91.95 $\pm$ 0.29 & 91.95 $\pm$ 0.29 & 92.14 $\pm$ 0.23 & 91.95 $\pm$ 0.29 & 91.95 $\pm$ 0.29 & 0.05 $\pm$ 0.00 & 92.04 $\pm$ 0.25 & 92.04 $\pm$ 0.25 & 92.17 $\pm$ 0.24 & 92.04 $\pm$ 0.25 & 92.04 $\pm$ 0.25 & 0.09 $\pm$ 0.00 \\ 
1 $\times$ [Bi]LSTM & 2 $\times$ [Bi]LSTM & 1 $\times$ [Bi]LSTM & 92.02 $\pm$ 0.29 & 92.02 $\pm$ 0.29 & 92.17 $\pm$ 0.24 & 92.02 $\pm$ 0.29 & 92.02 $\pm$ 0.29 & 0.07 $\pm$ 0.00 & 92.17 $\pm$ 0.26 & 92.17 $\pm$ 0.26 & 92.36 $\pm$ 0.23 & 92.17 $\pm$ 0.26 & 92.17 $\pm$ 0.26 & 0.11 $\pm$ 0.00 \\ 
1 $\times$ [Bi]LSTM & 2 $\times$ [Bi]LSTM & 2 $\times$ [Bi]LSTM & 91.87 $\pm$ 0.24 & 91.87 $\pm$ 0.24 & 91.98 $\pm$ 0.26 & 91.87 $\pm$ 0.24 & 91.87 $\pm$ 0.24 & 0.09 $\pm$ 0.00 & 91.94 $\pm$ 0.23 & 91.94 $\pm$ 0.23 & 92.03 $\pm$ 0.25 & 91.94 $\pm$ 0.23 & 91.94 $\pm$ 0.23 & 0.15 $\pm$ 0.00 \\ 
1 $\times$ [Bi]LSTM & 2 $\times$ [Bi]LSTM & 3 $\times$ [Bi]LSTM & 91.56 $\pm$ 0.32 & 91.56 $\pm$ 0.32 & 91.72 $\pm$ 0.32 & 91.56 $\pm$ 0.32 & 91.56 $\pm$ 0.32 & 0.11 $\pm$ 0.00 & 91.62 $\pm$ 0.27 & 91.62 $\pm$ 0.27 & 91.76 $\pm$ 0.27 & 91.62 $\pm$ 0.27 & 91.62 $\pm$ 0.27 & 0.18 $\pm$ 0.00 \\ 
1 $\times$ [Bi]LSTM & 3 $\times$ [Bi]LSTM & 1 $\times$ [Bi]LSTM & 91.86 $\pm$ 0.30 & 91.86 $\pm$ 0.30 & 92.01 $\pm$ 0.31 & 91.86 $\pm$ 0.30 & 91.86 $\pm$ 0.30 & 0.13 $\pm$ 0.00 & 91.97 $\pm$ 0.18 & 91.97 $\pm$ 0.18 & 92.09 $\pm$ 0.18 & 91.97 $\pm$ 0.18 & 91.97 $\pm$ 0.18 & 0.21 $\pm$ 0.00 \\ 
1 $\times$ [Bi]LSTM & 3 $\times$ [Bi]LSTM & 2 $\times$ [Bi]LSTM & 91.58 $\pm$ 0.52 & 91.58 $\pm$ 0.52 & 91.74 $\pm$ 0.51 & 91.58 $\pm$ 0.52 & 91.58 $\pm$ 0.52 & 0.15 $\pm$ 0.00 & 91.53 $\pm$ 0.28 & 91.53 $\pm$ 0.28 & 91.65 $\pm$ 0.33 & 91.53 $\pm$ 0.28 & 91.53 $\pm$ 0.28 & 0.25 $\pm$ 0.00 \\ 

1 $\times$ [Bi]LSTM & 3 $\times$ [Bi]LSTM & 3 $\times$ [Bi]LSTM & 91.09 $\pm$ 0.43 & 91.09 $\pm$ 0.43 & 91.21 $\pm$ 0.42 & 91.09 $\pm$ 0.43 & 91.09 $\pm$ 0.43 & 0.17 $\pm$ 0.00 & 91.07 $\pm$ 0.32 & 91.07 $\pm$ 0.32 & 91.17 $\pm$ 0.34 & 91.07 $\pm$ 0.32 & 91.07 $\pm$ 0.32 & 0.29 $\pm$ 0.00 \\ 
2 $\times$ [Bi]LSTM & 1 $\times$ [Bi]LSTM & 1 $\times$ [Bi]LSTM & \textbf{92.50} $\pm$ 0.38 & 92.50 $\pm$ 0.38 & 92.67 $\pm$ 0.35 & 92.50 $\pm$ 0.38 & 92.50 $\pm$ 0.38 & 0.19 $\pm$ 0.00 & \textbf{92.50} $\pm$ 0.26 & 92.50 $\pm$ 0.26 & 92.69 $\pm$ 0.21 & 92.50 $\pm$ 0.26 & 92.50 $\pm$ 0.26 & 0.32 $\pm$ 0.00 \\ 
2 $\times$ [Bi]LSTM & 1 $\times$ [Bi]LSTM & 2 $\times$ [Bi]LSTM & 92.30 $\pm$ 0.28 & 92.30 $\pm$ 0.28 & 92.46 $\pm$ 0.25 & 92.30 $\pm$ 0.28 & 92.30 $\pm$ 0.28 & 0.21 $\pm$ 0.01 & 92.35 $\pm$ 0.26 & 92.35 $\pm$ 0.26 & 92.52 $\pm$ 0.22 & 92.35 $\pm$ 0.26 & 92.35 $\pm$ 0.26 & 0.35 $\pm$ 0.01 \\ 
2 $\times$ [Bi]LSTM & 1 $\times$ [Bi]LSTM & 3 $\times$ [Bi]LSTM & 92.04 $\pm$ 0.37 & 92.04 $\pm$ 0.37 & 92.20 $\pm$ 0.38 & 92.04 $\pm$ 0.37 & 92.04 $\pm$ 0.37 & 0.24 $\pm$ 0.00 & 92.11 $\pm$ 0.25 & 92.11 $\pm$ 0.25 & 92.20 $\pm$ 0.21 & 92.11 $\pm$ 0.25 & 92.11 $\pm$ 0.25 & 0.39 $\pm$ 0.00 \\ 
2 $\times$ [Bi]LSTM & 2 $\times$ [Bi]LSTM & 1 $\times$ [Bi]LSTM & 92.36 $\pm$ 0.31 & 92.36 $\pm$ 0.31 & 92.48 $\pm$ 0.25 & 92.36 $\pm$ 0.31 & 92.36 $\pm$ 0.31 & 0.26 $\pm$ 0.00 & 92.40 $\pm$ 0.20 & 92.40 $\pm$ 0.20 & 92.52 $\pm$ 0.21 & 92.40 $\pm$ 0.20 & 92.40 $\pm$ 0.20 & 0.42 $\pm$ 0.01 \\ 
2 $\times$ [Bi]LSTM & 2 $\times$ [Bi]LSTM & 2 $\times$ [Bi]LSTM & 92.05 $\pm$ 0.23 & 92.05 $\pm$ 0.23 & 92.17 $\pm$ 0.21 & 92.05 $\pm$ 0.23 & 92.05 $\pm$ 0.23 & 0.28 $\pm$ 0.00 & 92.19 $\pm$ 0.25 & 92.19 $\pm$ 0.25 & 92.33 $\pm$ 0.22 & 92.19 $\pm$ 0.25 & 92.19 $\pm$ 0.25 & 0.46 $\pm$ 0.01 \\ 
2 $\times$ [Bi]LSTM & 2 $\times$ [Bi]LSTM & 3 $\times$ [Bi]LSTM & 91.70 $\pm$ 0.47 & 91.70 $\pm$ 0.47 & 91.87 $\pm$ 0.45 & 91.70 $\pm$ 0.47 & 91.70 $\pm$ 0.47 & 0.31 $\pm$ 0.01 & 91.69 $\pm$ 0.35 & 91.69 $\pm$ 0.35 & 91.88 $\pm$ 0.40 & 91.69 $\pm$ 0.35 & 91.69 $\pm$ 0.35 & 0.51 $\pm$ 0.01 \\ 
2 $\times$ [Bi]LSTM & 3 $\times$ [Bi]LSTM & 1 $\times$ [Bi]LSTM & 92.13 $\pm$ 0.28 & 92.13 $\pm$ 0.28 & 92.36 $\pm$ 0.27 & 92.13 $\pm$ 0.28 & 92.13 $\pm$ 0.28 & 0.33 $\pm$ 0.01 & 92.16 $\pm$ 0.28 & 92.16 $\pm$ 0.28 & 92.33 $\pm$ 0.27 & 92.16 $\pm$ 0.28 & 92.16 $\pm$ 0.28 & 0.54 $\pm$ 0.01 \\ 
2 $\times$ [Bi]LSTM & 3 $\times$ [Bi]LSTM & 2 $\times$ [Bi]LSTM & 91.94 $\pm$ 0.32 & 91.94 $\pm$ 0.32 & 92.06 $\pm$ 0.33 & 91.94 $\pm$ 0.32 & 91.94 $\pm$ 0.32 & 0.35 $\pm$ 0.01 & 91.69 $\pm$ 0.32 & 91.69 $\pm$ 0.32 & 91.79 $\pm$ 0.37 & 91.69 $\pm$ 0.32 & 91.69 $\pm$ 0.32 & 0.58 $\pm$ 0.01 \\ 
2 $\times$ [Bi]LSTM & 3 $\times$ [Bi]LSTM & 3 $\times$ [Bi]LSTM & 91.18 $\pm$ 0.25 & 91.18 $\pm$ 0.25 & 91.30 $\pm$ 0.27 & 91.18 $\pm$ 0.25 & 91.18 $\pm$ 0.25 & 0.38 $\pm$ 0.01 & 91.18 $\pm$ 0.23 & 91.18 $\pm$ 0.23 & 91.29 $\pm$ 0.24 & 91.18 $\pm$ 0.23 & 91.18 $\pm$ 0.23 & 0.63 $\pm$ 0.01 \\ 
3 $\times$ [Bi]LSTM & 1 $\times$ [Bi]LSTM & 1 $\times$ [Bi]LSTM & 92.38 $\pm$ 0.25 & 92.38 $\pm$ 0.25 & 92.50 $\pm$ 0.22 & 92.38 $\pm$ 0.25 & 92.38 $\pm$ 0.25 & 0.40 $\pm$ 0.01 & 92.34 $\pm$ 0.21 & 92.34 $\pm$ 0.21 & 92.45 $\pm$ 0.23 & 92.34 $\pm$ 0.21 & 92.34 $\pm$ 0.21 & 0.67 $\pm$ 0.01 \\ 
3 $\times$ [Bi]LSTM & 1 $\times$ [Bi]LSTM & 2 $\times$ [Bi]LSTM & 92.13 $\pm$ 0.30 & 92.13 $\pm$ 0.30 & 92.23 $\pm$ 0.30 & 92.13 $\pm$ 0.30 & 92.13 $\pm$ 0.30 & 0.43 $\pm$ 0.01 & 92.16 $\pm$ 0.27 & 92.16 $\pm$ 0.27 & 92.26 $\pm$ 0.26 & 92.16 $\pm$ 0.27 & 92.16 $\pm$ 0.27 & 0.71 $\pm$ 0.01 \\ 
3 $\times$ [Bi]LSTM & 1 $\times$ [Bi]LSTM & 3 $\times$ [Bi]LSTM & 91.87 $\pm$ 0.26 & 91.87 $\pm$ 0.26 & 92.13 $\pm$ 0.24 & 91.87 $\pm$ 0.26 & 91.87 $\pm$ 0.26 & 0.45 $\pm$ 0.01 & 92.06 $\pm$ 0.23 & 92.06 $\pm$ 0.23 & 92.16 $\pm$ 0.26 & 92.06 $\pm$ 0.23 & 92.06 $\pm$ 0.23 & 0.76 $\pm$ 0.01 \\ 
3 $\times$ [Bi]LSTM & 2 $\times$ [Bi]LSTM & 1 $\times$ [Bi]LSTM & 92.28 $\pm$ 0.28 & 92.28 $\pm$ 0.28 & 92.42 $\pm$ 0.26 & 92.28 $\pm$ 0.28 & 92.28 $\pm$ 0.28 & 0.48 $\pm$ 0.01 & 92.22 $\pm$ 0.17 & 92.22 $\pm$ 0.17 & 92.37 $\pm$ 0.11 & 92.22 $\pm$ 0.17 & 92.22 $\pm$ 0.17 & 0.80 $\pm$ 0.01 \\ 
3 $\times$ [Bi]LSTM & 2 $\times$ [Bi]LSTM & 2 $\times$ [Bi]LSTM & 92.04 $\pm$ 0.25 & 92.04 $\pm$ 0.25 & 92.20 $\pm$ 0.31 & 92.04 $\pm$ 0.25 & 92.04 $\pm$ 0.25 & 0.50 $\pm$ 0.01 & 92.04 $\pm$ 0.25 & 92.04 $\pm$ 0.25 & 92.17 $\pm$ 0.26 & 92.04 $\pm$ 0.25 & 92.04 $\pm$ 0.25 & 0.84 $\pm$ 0.01 \\ 
3 $\times$ [Bi]LSTM & 2 $\times$ [Bi]LSTM & 3 $\times$ [Bi]LSTM & 91.14 $\pm$ 0.21 & 91.14 $\pm$ 0.21 & 91.19 $\pm$ 0.24 & 91.14 $\pm$ 0.21 & 91.14 $\pm$ 0.21 & 0.52 $\pm$ 0.01 & 91.30 $\pm$ 0.11 & 91.30 $\pm$ 0.11 & 91.43 $\pm$ 0.12 & 91.30 $\pm$ 0.11 & 91.30 $\pm$ 0.11 & 0.88 $\pm$ 0.01 \\ 
3 $\times$ [Bi]LSTM & 3 $\times$ [Bi]LSTM & 1 $\times$ [Bi]LSTM & 92.19 $\pm$ 0.23 & 92.19 $\pm$ 0.23 & 92.31 $\pm$ 0.25 & 92.19 $\pm$ 0.23 & 92.19 $\pm$ 0.23 & 0.55 $\pm$ 0.01 & 92.19 $\pm$ 0.18 & 92.19 $\pm$ 0.18 & 92.27 $\pm$ 0.21 & 92.19 $\pm$ 0.18 & 92.19 $\pm$ 0.18 & 0.92 $\pm$ 0.02 \\ 
3 $\times$ [Bi]LSTM & 3 $\times$ [Bi]LSTM & 2 $\times$ [Bi]LSTM & 92.10 $\pm$ 0.13 & 92.10 $\pm$ 0.13 & 92.19 $\pm$ 0.12 & 92.10 $\pm$ 0.13 & 92.10 $\pm$ 0.13 & 0.58 $\pm$ 0.02 & 91.57 $\pm$ 0.32 & 91.57 $\pm$ 0.32 & 91.70 $\pm$ 0.35 & 91.57 $\pm$ 0.32 & 91.57 $\pm$ 0.32 & 0.97 $\pm$ 0.02 \\ 
3 $\times$ [Bi]LSTM & 3 $\times$ [Bi]LSTM & 3 $\times$ [Bi]LSTM & 91.01 $\pm$ 0.09 & 91.01 $\pm$ 0.09 & 91.09 $\pm$ 0.08 & 91.01 $\pm$ 0.09 & 91.01 $\pm$ 0.09 & 0.60 $\pm$ 0.02 & 91.03 $\pm$ 0.17 & 91.03 $\pm$ 0.17 & 91.14 $\pm$ 0.20 & 91.03 $\pm$ 0.17 & 91.03 $\pm$ 0.17 & 1.02 $\pm$ 0.02 \\ 
\hline
		\end{tabular}}
\end{table}

\section{Discussion}\label{sec:discussion}

The misinformation detection problem is a complex task that requires large datasets to accurately determine news content veracity.
We found that in the current literature, many authors utilize relatively small datasets for this task, e.g., datasets with $\sim$1,000
articles in~\cite{Kurasinski2020}, $\sim$10,000 
in~\cite{Gautam2021}, which is negatively reflected in the accuracy obtained by the tested models, especially when it comes to BERT-based models.
Surely, there are authors who test their solutions using larger corpora, e.g.,~$\sim$80,000 articles in~\cite{Khan2021benchmark}.
In some cases, the dataset consists of short news statements, which in turn affect the accuracy considerably~\cite{Liu2019BertFN}.
Furthermore, the current research articles only address the misinformation detection problem using either a binary approach (the news article's veracity is either true or false~\cite{Khan2021benchmark,Kaliyar2021,Kurasinski2020}) or levels of veracity~\cite{Kula2020}.
Thus, none of the current approaches uses a large dataset that contains fake news labeled with different categories.
We address this shortcoming by using a large dataset containing 100,000 news articles labeled with 10~classes.

The size of the corpus, the number of tokens, the used vocabulary, the frequent unigrams, and topics play an important role in the overall and per-class detection task.
Through fine-tuning, we include these dimensions in our models, thus improving the classification performance.

In the misinformation detection task, contextual information plays an important role that impacts the accuracy of determining hidden patterns within the text.
Thus, using transformers and transfer learning, the models better discriminate between different types of misinformation as shown by the recall score of over 85\%.
By adding autoregression, sequence-to-sequence with a bidirectional encoder, and a left-right decoder, BART large achieves the overall best performance among the basic architecture with transformers.

For BERT, RoBERTa, DeBERTa, XLNet, ELECTRA, and XLM-RoBERTa, the base models manage to better generalize than the large ones, while for BART and ALBERT, the reverse happens.
Thus, we can conclude that BERT, RoBERTa, DeBERTa, XLNet, ELECTRA, and XLM-RoBERTa manage to extract correctly the hidden context within the misinformation texts and better discriminate between the labels with a smaller network size and number of parameters.
The autoregressive method to learn bidirectional contexts employed by XLNet and BART improves the misinformation detection performance as the network and number of parameters increases. 
We can conclude that, for XLNet base, although extracting less information from context than large, it achieves better results through the autoregressive method.
Thus, both XLNet base and BART (base and large) extract more information from context using the autoregressive~method.

XLM, a multilingual model, performs rather poorly.
This is somehow expected as the best performing multilingual system scored below the worst performing native system~\cite{Zampieri2020}.

Although distillation is a good technique to maintain classification accuracy by lowering resource consumption, DistilBERT and ALBERT models do not provide better results than the base models.
As ALBERT also reduces BERT's number of parameters, it still needs a large network to achieve good results. 
Furthermore, ALBERT performance significantly improves when dropout is used, which also lowers execution performance.
These models can still be used in low resource environments, offering an adequate level of accuracy.
We can conclude that by reducing both the number of parameters and the network size, the classification performance is decreased for these models. 

DistilRoBERTa achieves the best accuracy (91.32\%) among the distilled versions and the overall smallest runtime (1.93h for fine-tuning and training).
Its accuracy falls short with only 0.63\% than the model with the highest accuracy, BART large.
In conclusion, DistilRoBERTa model can be a viable option when the number of resources is limited, and~both fine-tuning and training should be conducted with the least amount of time. 

MisRoBÆRTa is a BiLSTM-CNN deep learning architecture that utilizes as input both RoBERTa and BART and sentence embeddings.
This novel architecture manages to better preserve context than the transformer models, as it employs context-aware and future selection layers besides context-aware embeddings.
MisRoBÆRTa outperforms the BART large model w.r.t. classification performance and DistilRoBERTa model w.r.t. execution~time.

Finally, we would like to conclude the discussion of our results by answering one final question: \textit{is misinformation detection a solved problem?}
In short, no.
Although the presented transformer-based methods and the novel architecture MisRoBÆRTa show their effectiveness by obtaining high accuracy using linguistic features, there remain a lot of challenges that, if~tackled, will improve performance and help users make informed decisions about the content of their news feeds.
Some of the challenges we identified while surveying the related work for the task of misinformation detection that requires a shift from a model-centric perspective to a data-centric perspective are:
\begin{itemize}
    \item[(\textit{1})] Multi-modal (i.e., visual, audio, network propagation, and immunization, etc.) techniques are rarely used; this is due to a lack of complete, high quality, and~well labeled corpora and the tedious work that is required to correctly annotate such datasets;
    \item[(\textit{2})] Source verification is rarely taken into consideration for misinformation detection~\cite{parikh2018media}, this is due to the instability of links on the internet;
    \item[(\textit{3})] Author credibility should be an important factor that is not really discussed or weighted in the feature selection for the models;
    \item[(\textit{4})] The perpetual need to retrain or fine-tune models to capture the subtle shifts that appear over time in news articles that spread misinformation.
\end{itemize}

\section{Conclusions}\label{sec:conclusions}

In this paper, we present a performance evaluation of transformers on the task of misinformation detection.
We employed a large real-world dataset for our experiments and addressed two shortcomings in the current research: 
(\textit{1}) Increasing the size of the dataset from small to large; and~
(\textit{2}) Moving the focus of fake news detection from binary to multi-class classification.
We present and discuss our findings in detail.
Furthermore, the results show that the accuracy of transformers on the misinformation detection task is significantly influenced by the method employed to learn the context, dataset size, and vocabulary.

We propose MisRoBÆRTa, a new transformer-based deep learning architecture for misinformation detection.
The proposed architecture used, as input, BART and RoBERTa sentence embeddings and a BiLSTM-CNN architecture for each embedding.
The output of each BiLSTM-CNN was concatenated and passed through another BiLSTM-CNN to obtain the results. We conclude that the proposed architecture outperforms the other employed classification models w.r.t. (\textit{1}) classification performance, obtaining an average accuracy of 92.50\%; and~(\textit{2}) runtime, with an average execution time for training of 1.83h.

We respond to the research questions as follows.
The overall best accuracy among the transformer models is obtained by BART (91.94\%), while DistilRoBERTa obtains the best accuracy (91.32\%) in the least amount of time required for fine-tuning and training, i.e., 1.93h on average.
MisRoBÆRTa outperforms both of these models' w.r.t. accuracy (92.50\%) and performance times (1.83h).
We empirically observed  that between BART base and large, the difference in performance is insignificant, while the difference in runtime is of almost 2.62h on average.
Moreover, some improved models of BERT slightly outperform the base model, i.e., DeBERTa and RoBERTa, while most fall short.
Furthermore, multilingual models, such as XLM, do not obtain better accuracy than BERT base.

In future research, our solution can be enhanced with sentiment analysis techniques that will determine the polarity and strength of the sentiments expressed in each news article.
Another research direction is to add global context through topic modeling~\cite{Mitroi2020,Truica2021}.

\paragraph*{Acknowledgments}
The publication of this paper is supported by the University Politehnica of Bucharest through the PubArt program and the work has been partly funded by EU CEF grant number 2394203 (NORDIS - NORdic observatory for digital media and information DISorders).

\end{document}